# DN-Hypo-Pipeline: An AI-Driven Workflow for Generating Hypotheses using Large Language Models and Scientific Explanations

Lei LIN, linlei@cnic.cn, Computer Network Information Center, Chinese Academy of Sciences, China.
Xinlong PAN, airadar@126.com, National Key Laboratory of Millimeter-Wave and Terahertz Remote Sensing, Naval Aviation University, Yantai, 264001, China.
Ronghao WANG, rhwang@cnic.cn, Computer Network Information Center, Chinese Academy of Sciences, China.
Chunbao ZHOU, zhoucb@sccas.cn, Computer Network Information Center, Chinese Academy of Sciences, China.
Jue WANG, wangjue@sccas.cn, Computer Network Information Center, Chinese Academy of Sciences, China.
Yangang WANG, wangyg@sccas.cn, Computer Network Information Center, Chinese Academy of Sciences, China.
Ivana RASOVSKA, ivana.rasovska@unistra.fr, ICube UMR 7357 Laboratory, University of Strasbourg, France.

**Abstract:** Modern AI systems are powerful predictors but remain limited in producing principled scientific explanations. We ask whether the structure of scientific explanation can be operationalized to guide hypothesis generation. We introduce DN-Hypo-Pipeline, a framework built on an explanation-theoretic scaffold: Hempel's deductive-nomological model, Salmon's causal-process account, and Armstrong's view of laws as relations between universals. Given an explanandum, the framework abstracts universals from the phenomenon's formation process, retrieves the laws relating those universals, and deductively reconstructs a testable explanation. Evaluated in data-science modeling and judged by both LLMs and human experts, hypotheses generated through this principled reasoning significantly outperform direct prompting. We further implemented the two highest-scoring hypotheses as algorithms: one reduces Transformer theoretical complexity with minimal performance loss, and another achieves competitive word-embedding accuracy with substantially fewer parameters. These results provide an initial demonstration that explanation-theoretic structure can support computational hypothesis generation, while broader cross-domain validation remains necessary.

**Keywords:** machine reasoning, scientific explanation, AI for science, deductive-nomological model, hypothesis generation

# 1. Introduction

**The prediction—explanation divide in machine intelligence.** Artificial intelligence has become extraordinarily good at one thing: prediction. Large language models predict the next token; AI-for-science systems predict structures, properties, and outcomes with superhuman accuracy. Yet beneath this success lies a persistent limitation that the scientific community has only begun to confront. These systems are, at their core, engines of association: they tell us what is likely, by interpolating over patterns already present in the human record. They do not tell us why. And it is precisely the why—the explanation of a phenomenon in terms of the principles that govern it—that has always been the deepest aim of science (*1*, *2*). A model that predicts the orbit of a planet without invoking gravitation is, in a scientific sense, incomplete. To genuinely model a phenomenon is to explain it.

**What a scientific explanation is.** This intuition was made precise by Hempel (*3*), whose deductive-nomological (DN) model holds that to explain a phenomenon is to show that its occurrence follows, as a matter of logical necessity, from general laws together with the particular conditions at hand. Consider photosynthesis: we do not explain it merely by reporting that plants turn sunlight into energy, but by appealing to the law-governed mechanisms—the light-dependent reactions, the Calvin cyclethrough which light energy and carbon dioxide are necessarily transformed into chemical energy. A scientific hypothesis, in this view, is a candidate explanation: a conjecture, grounded in existing knowledge, about the laws and conditions that would account for an observed phenomenon, formulated so as to be testable (*1*, *4*). Generating a good hypothesis is therefore not an act of recall but an act of reasoning—and it is among the hardest and most consequential steps in all of science.

**How today's AI generates hypotheses—and why it falls short.** A growing body of work now enlists LLMs across the research pipeline, from idea generation and experimental design to analysis (*5*–*7*), and in some domains these systems already rival human scientists in narrow inferential tasks (*8*, *9*). Yet almost all of these efforts share a common architecture and a common ceiling. They generate ideas the way a busy researcher might: by reviewing the existing literature and recombining what has been reported (*10*). This is a powerful strategy, but a fundamentally bounded one. A system whose hypothesis space is defined by what has already been written can, at best, interpolate within human knowledge; it cannot reason outward to an explanation that the literature does not yet contain. More tellingly, these systems imitate the observable behavior of a scientist—the workflow of searching, reading, and proposing—without ever formalizing the reasoning engine that makes scientific discovery possible in the first place. They model the scientist's actions, not the scientist's logic.

**Our proposal: operationalizing the structure of explanation.** We take a different starting point. Rather than asking a machine to mimic what scientists do, we ask whether the structure of scientific explanation can be operationalized to organize how a machine generates hypotheses. We introduce DN-Hypo-Pipeline, in which three classical resources from the philosophy of explanation are assigned distinct, complementary roles rather than treated as competing theories. Hempel's DN model defines the output form and deductive validity of a hypothesis: a candidate explanation must subsume the explanandum under general laws together with initial conditions, such that the explanandum follows as a logical consequence. The DN covering-law criterion, however, is by itself too permissiveit cannot, on its own, separate genuinely explanatory laws from accidental or non-causal correlations. To narrow the search and exclude such spurious associations, we adopt Salmon's causal-process (*11*, *12*) account as an organizing constraint on the search space: the universals admitted into a candidate explanation must be those instantiated by the causal processes actually present in the phenomenon's formation, not abstract regularities introduced after the fact. Armstrong's thesis (*13*) that laws are relations between universals then provides the bridge from process to law: once the universals in the formation process are identified, the problem of finding the governing laws becomes the problem of retrieving the laws that relate those universals. Within this layered design, the LLM serves as the interpretable articulator that instantiates each step. The shift is consequential: instead of searching the space of what has been written, the framework searches the space of what principles govern a phenomenon, and the hypothesis space is determined not by the expressiveness of a template or the contents of a corpus, but by the structural requirements of explanation. We emphasize that this layering is a deliberate methodological design in service of engineering feasibility, and not a claim to have unified these accounts at the philosophical level—a limitation we return to in Section 6.

**Why this matters beyond data science.** Because the framework reasons from the structure of explanation rather than from a domain-specific literature, it is not tied to any one field. It is, in effect, a generalization of theory-guided modeling (*14*) — an attempt to expose and operationalize the common logical skeleton that underlies how phenomena are explained across disciplines. We demonstrate the framework in the field of data-science modeling, where ground truth is accessible and hypotheses can be turned into running algorithms and tested directly; but the same machinery, in principle, applies wherever phenomena demand explanation, from the natural to the social sciences.

**Contributions.** Our contributions are threefold. First, we present DN-Hypo-Pipeline, a hypothesis-generation framework that makes a preliminary yet principled attempt to operationalize the structure of scientific explanation, rather than to imitate scientific output empirically or heuristically. Concretely, it organizes the hypothesis space through a layered explanation-theoretic design—the DN model fixing the deductive output form, Salmon's causal-process account constraining the search for governing laws, and Armstrong's universals bridging from a phenomenon's formation process to those laws thereby offering a principled, if still early-stage, basis for modeling where and how candidate explanations are sought. Second, through a comprehensive statistical evaluation judged by both LLMs and human experts, we show that hypotheses generated through principled reasoning significantly outperform those produced by direct prompting; We further benchmark a suite of LLMs within the framework, identifying GPT-5.2 as the strongest performer. Under the Manhattan distance metric, we find that—apart from Gemini-3.1-Pro-Preview, which aligns with only one of the three human experts—the LLM-as-a-judge assessments closely mirror the judgments of the other two human experts. Third, we close the loop from explanation to evidence: we translate the two highest-scoring hypotheses into novel, working algorithms—one that theoretically reduces the computational complexity of the Transformer with only minimal performance degradation, and another that vectorizes words by isolating a frequency-adaptive expression space—each of which outperforms the baseline reported in the original paper.

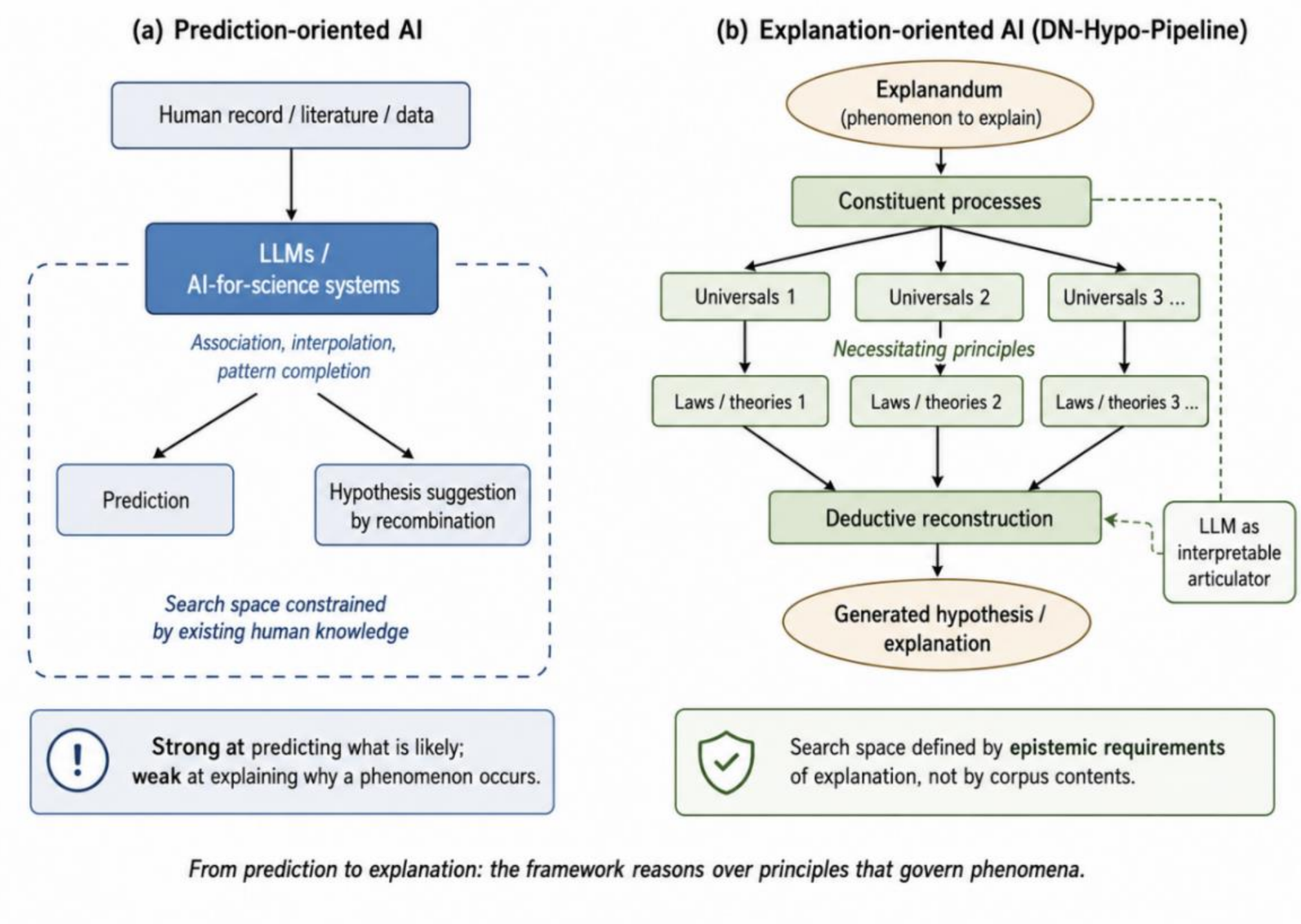

Fig. 1. From prediction-oriented AI to explanation-oriented hypothesis generation

# 2. Related work

Recent work that uses large language models (LLMs) to generate scientific hypotheses can be characterized along three axes: (a) how a hypothesis is represented, (b) what external knowledge grounds it, and (c) how the generation process is modeled. Despite this diversity, the literature consistently treats hypothesis generation as a problem that requires constructing a text-, graph-, or reasoning-chain construction problem. The design choices are overwhelmingly guided by pragmatic criteria—the structure of a paper, the format of a dataset, or the mechanics of an agent loop—rather than by an epistemological analysis of what a scientific hypothesis needs to satisfy. This omission has produced a collection of heuristically motivated systems that, while individually productive, operate in spaces whose boundaries are never formally justified.

**Hypothesis representations: from structured output formats to knowledge-graph links.** Systems cluster around two broad representational strategies, with no intended progression between them. The first strategy imposes a structured textual schema on the output: in-context or adversarial prompting produces unstructured proposals or cross-domain concept associations (*15–17*), while more explicitly structured decompositions use purpose-mechanism-evaluation triples (*18*), problem-method-experiment designs (*19*), Bit-Flip-Spark chains (*20*), or method-experiment plans extracted from papers (*21*). The second strategy abandons natural-language prose in favor of knowledge-graph links, representing hypotheses as causal edges between genes (*22*), key research aspects (e.g., hypothesis, mechanisms, and design principles) synthesized from subgraph paths sampled via heuristic pathfinding (*23*), or entity-entity causal labels decorated with reasoning chains (*24*). These two approaches differ in their medium, yet they share a common origin: in every case, the representation is derived from the surface structure of academic papers, curated datasets, or domain ontologies. The hypothesis space is thus circumscribed by the expressiveness of the chosen template, not by the epistemic demands of a scientific hypothesis such as falsifiability, explanatory depth, or principle consistency. None of the cited works justify their representational choice from these first-principle requirements.

**Knowledge sources: bibliometric priors vs. principle-level grounding.** External knowledge is supplied through three channels—first, curated bibliometric corpora-paper abstracts, citation graphs, background-hypothesis pairs, or review scores-drive in-context retrieval, fine-tuning, and iterative novelty boosting (*15*, *17*, *18*, *21*, *25*–*27*). Second, structured knowledge bases (gene databases, large-scale ontologies, PubTator3 KG) provide entity-centric priors for graph-based methods (*22*–*24*). Third, a few ambitious frameworks augment these static sources with active retrieval loops: Co-Scientist (*6*) performs web searches and utilizes domain-specific open databases, together with specialized AI models (e.g., AlphaFold), while AI Scientist (*7*) leverages Semantic Scholar plus code execution logs to filter and cite. Despite their complexity, all these pipelines operate on "what has been reported" rather than "what principles govern the phenomenon". The resulting hypotheses tend to recombine known entities and relations, but struggle to generate explanations that are physically or biologically plausible yet absent from the literature. The gap becomes conspicuous when one contrasts these approaches with PiFlow (*28*) and PiEvo (*29*), which ground hypothesis generation not in the literature but in high-fidelity surrogate models (DFT/GNN) and a dynamically expanding set of physical principles. In that paradigm, the literature serves only as a weak prior to reduce the cost of brute-force exploration, and the quality of a hypothesis is measured by its ability to reduce uncertainty over the principle space.

**The generation process: from behavior mimicry to principle-aware inference.** The generation process ranges from simple direct prompting to multi-agent evolutionary loops. At the behavioral-mimicry end, models are fine-tuned to reproduce the reasoning patterns observed in existing papers: evidence chain prediction and validation (*30*), Bit-Flip-Spark extraction (*20*), idea enumeration with learned ranking (*25*), or controllable reinforcement learning that maps a paper to a future method-experiment plan (*21*). More elaborate frameworks decompose the process into inspiration retrieval, hypothesis composition, and iterative feedback (MOOSE (*27*), MOOSE-Chem (*31*), ResearchBench (*32*)), sometimes orchestrating a tournament of generation, reflection, ranking, and evolution (Co-Scientist (*6*)), or closing the loop with automated paper writing and reviewing (AI Scientist (*7*)). While these systems simulate increasingly rich scientific workflows, they nevertheless mimic the scientist's observable behavior without formalizing the inference engine that drives discovery. The search strategies remain heuristic ("prompt and select") or data-driven ("fine-tune on paper structures"), and the hypothesis space is never defined in terms of the underlying logic of abduction and falsification. This is precisely the distinction made explicit in PiEvo (*29*): it employs abductive reasoning to minimize uncertainty over a principle space, using Bayesian optimization with information-directed sampling and anomaly-driven principle evolution. The hypothesis is not a text string or a graph link but a set of principles together with specific physical parameters, evaluated by consistency with high-fidelity simulations. The entry point is the phenomenon-principle relationship, not the paper.

**The missing dimension: principled hypothesis space definition.** The literature thus exhibits a sharp divide. Nearly all LLM-based methods construct a pipeline that operationalizes some aspect of a scientist's workflow—reading, retrieving, composing, reviewing—while leaving the structure of the hypothesis space implicit in the chosen representation and data source. PiEvo and PiFlow demonstrate the alternative: a methodology-grounded approach in which the hypothesis space is explicitly defined by a set of principles, and generation is driven by principled inference rather than by imitation of scientific output. However, they are cost-prohibitive and require iterative experimentation or simulation. Nor can they exploit the wealth of domain knowledge encoded in the scientific literature beyond using it as a prior. Our work stands at this junction: we aim to integrate the principle-aware perspective into a literature-driven paradigm, equipping LLMs to generate hypotheses that are not only novel and literature-grounded, but also structurally aligned with the epistemological demands of a scientific theory.

In the following section, we introduce our proposed method, in which the representation of a scientific hypothesis is derived from its definition, and the generation process is formulated as a search for laws and the reconstruction of explanations for scientific phenomena.

# 3. Methodology

## 3.1 Theoretical Foundation & Methodological Logic

A scientific hypothesis is a proposed explanation for a phenomenon (*4*), while a scientific explanation is a structured account of why a phenomenon occurs that goes beyond mere description (*3*, *33*). Rather, it answers "why" questions regarding when and how a specific event or general law occurs, rather than merely stating "what" it is. Thus, we propose a pipeline that produces a structured scientific explanation for a phenomenon.

Among the scientific explanation models, the DN model is the most influential. This classic model, also called the covering law model, was developed by Carl G. Hempel and Paul Oppenheim (*34*) in 1948. According to this model, an explanation consists of:

| **An explanandum (E)** | A sentence describing the empirical phenomenon to be explained. |
|---|---|
| **The explanans** | A set of sentences that includes at least one **law of nature (L)** (i.e., a universal, law-like generalization), and some **initial conditions (C)** (i.e., some specific facts about the situation). |

The explanandum must be a logical consequence of the explanans in that the two combined must form a sound deductive argument. Further, the law of nature must be essential. As such, removing it would make the deduction invalid. This can be formally represented as:

| | | |
|---|---|---|
| **The law(s) of nature** | $L_1, L_2, L_3, \cdots, L_r$ | L and C create the logical deduction of E |
| **The initial conditions** | $C_1, C_2, C_3, \cdots, C_k$ | |
| **The explanandum** | **E** | |

Within the formal representation of the DN model, a scientific hypothesis is defined as a potential explanation that subsumes the explanandum under established general laws and specific initial conditions, such that one can deduce the explanandum, pending empirical confirmation.

Naturally, the hypothesis generation problem can be formulated as a mask generation problem. Hence, if we have a phenomenon (i.e., an explanandum), we are looking for a set of laws and conditions to deduce the phenomenon. This can be expressed as:

$$h \sim P(L_1, L_2, L_3 \cdots, C_1, C_2, C_3 \cdots | E) \quad (1)$$

Notably, Eq. (1) implies a probability generation model $P(\cdot \mid \cdot)$, such as an LLM that is given a background or previous context.

In the DN model, laws are the core components that supply the necessary connections between antecedent conditions and the phenomenon to be explained. Conversely, the initial conditions merely specify the particular circumstances to which those laws apply (*33*, *34*). Thus, the "initial conditions" are treated as supplementary content derived from generating "laws". From this standpoint, generating a hypothesis is a function of the following expression:

$$h \sim P(L_1, L_2, L_3 \cdots | E)$$

Moreover, the process of generating an explanandum can be represented as a thought-prompting problem, as outlined in Fig. 2(a). As shown, the key problem with generating a hypothesis is how to search through the laws that can potentially explain the phenomenon. Wesley Salmon (*11*, *12*, *33*) holds that the explanatory elements (i.e., the explanans or causal processes and interactions) must actually exist in the process of formulating the explanandum. Critically, they cannot merely be abstract laws or statistical correlations. Thus, one should be able to search for the associated laws in a structured space dedicated to forming an explanandum. The hypothesis generation problem can be expressed as:

$$h \sim P(L_1, L_2, L_3 \cdots | E, Processes) P(Processes | E) \quad (2)$$

where $Processes$ is the description of the formation process.

Further, Armstrong (*13*) argues that laws are simply relationships between universals. A universal is defined as a repeatable entity that can be wholly present in multiple distinct states of affairs. As such, universals can be associated with laws:

$$Law \sim P(L | \mathrm{u}) \quad (3)$$

where $L$ denotes the law, and $u$ denotes the universal. Thus, as described in Fig. 2(b), hypothesis generation can be further improved as follows:

$$h \sim P(L_1, L_2, L_3 \cdots | E, Process, u_1, u_2, u_3 \cdots) P(u_1, u_2, u_3 \cdots | E, Process) P(Process | E) \quad (4)$$

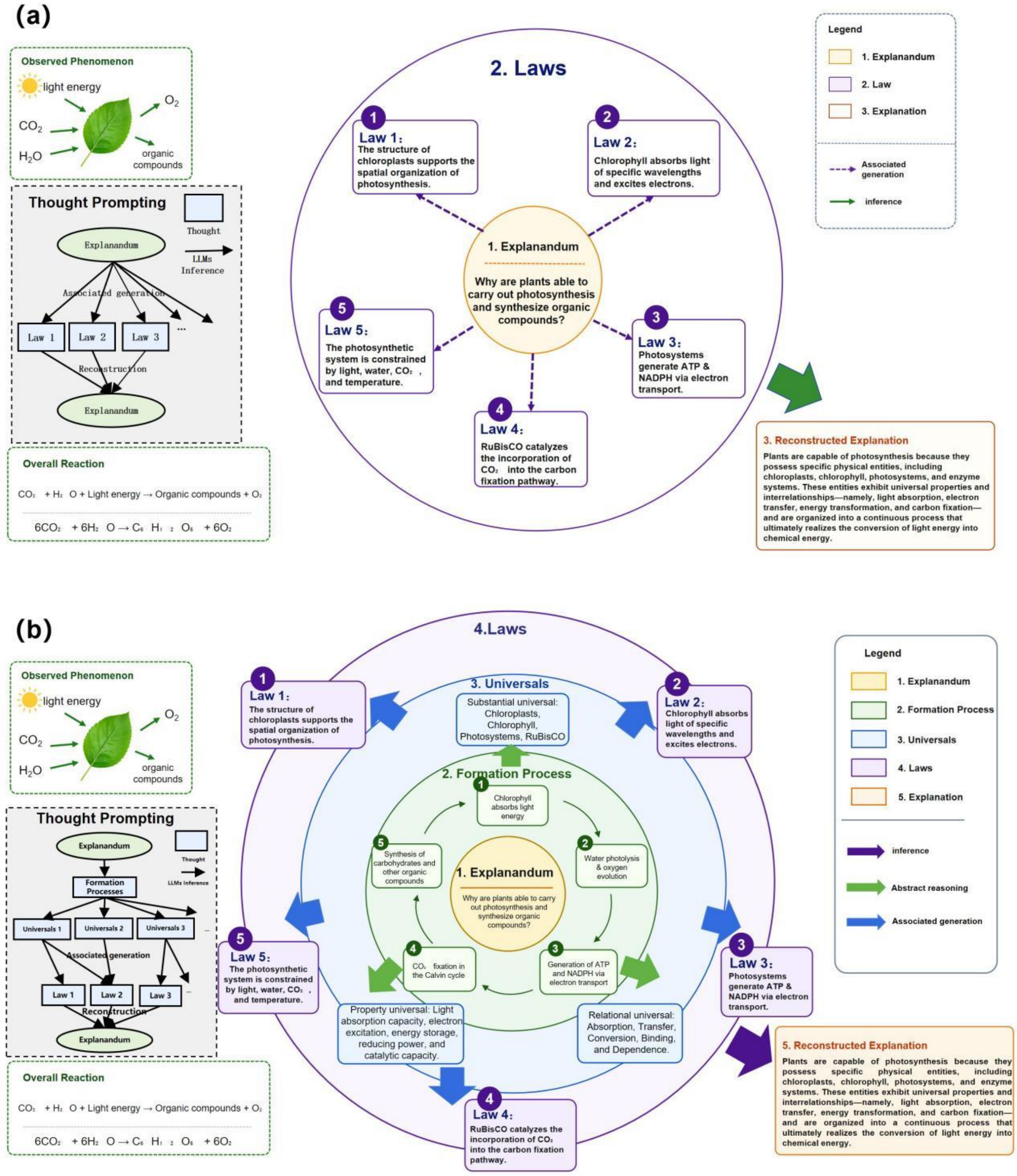

Fig. 2. Thought-prompting logic for hypothesis generation based on the deductive-nomological model. Photosynthesis is used as an illustrative example in both panels. (a) Thought-prompting logic for generating hypotheses by identifying laws and initial conditions that deductively explain an explanandum. (b) Extended thought-prompting logic for generating hypotheses by modeling the processes underlying the explanandum, identifying the universals instantiated in those processes, and associating the universals with relevant laws.

## 3.2 Methodological Framework

The proposed framework, DN-Hypo-Pipeline, generates hypotheses that identify potential research problems. As shown in Fig. 3(a), the system inputs a research paper and executes a hypothesis generation process. This process is enhanced by a corpus of research papers (via law retrieval) and a novelty checker, which ultimately delivers a list of hypotheses to the researcher.

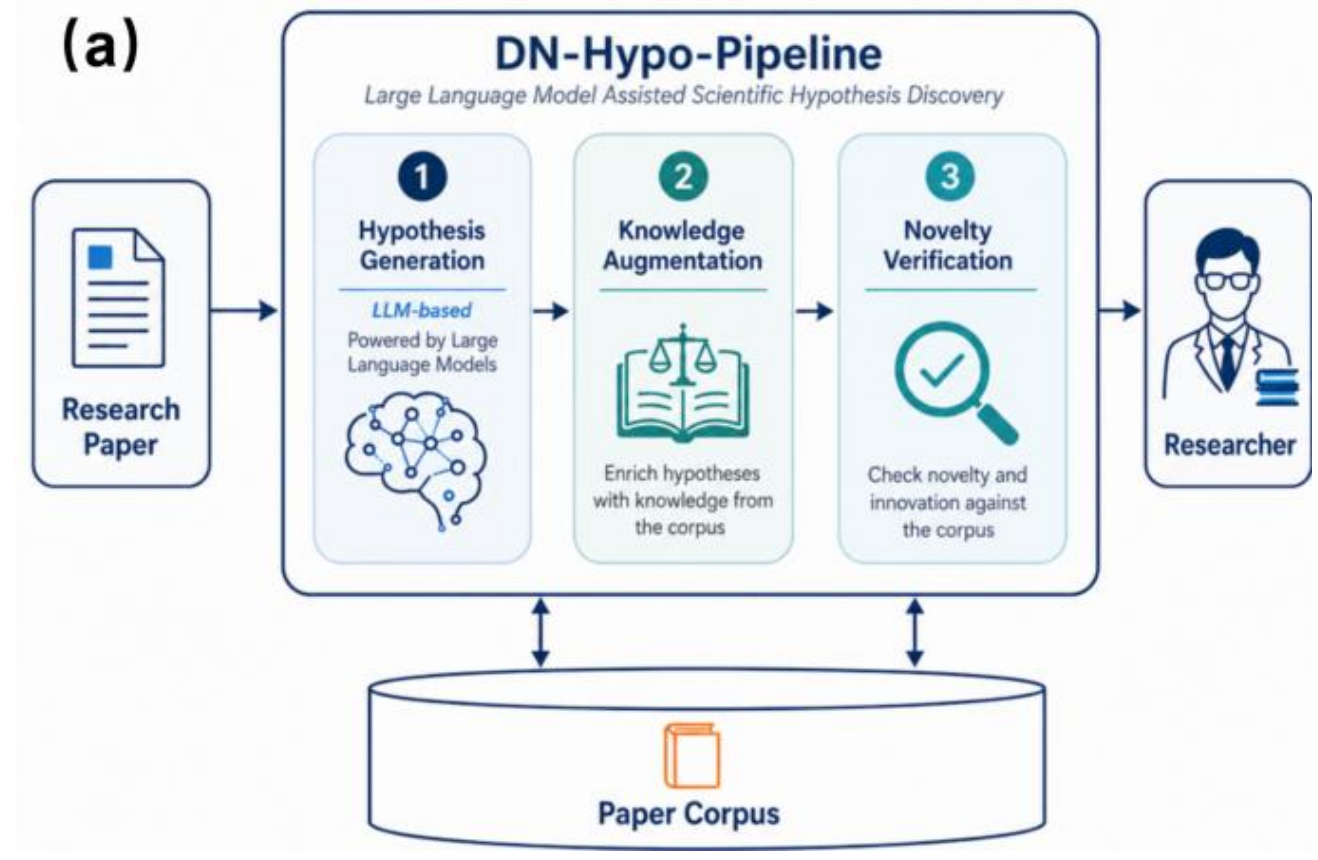

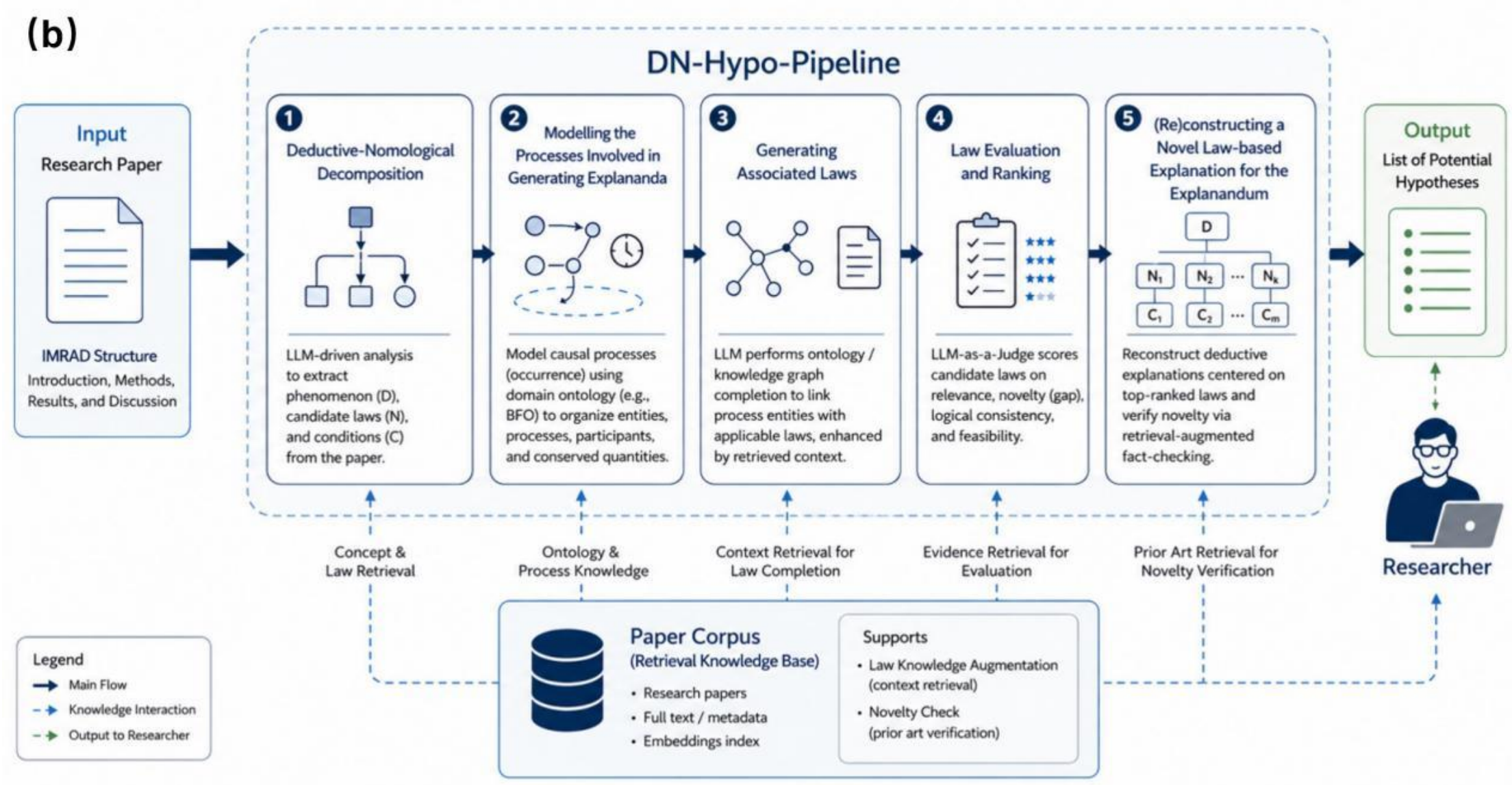

Fig. 3. Overview and detailed workflow of the DN-Hypo-Pipeline. (a) The overall framework of the DN-Hypo-Pipeline, illustrating hypothesis generation, knowledge augmentation, and novelty verification. (b) The detailed workflow, including Deductive-Nomological decomposition, causal-process modeling, associated-law generation, law evaluation and ranking, and reconstruction of novel law-based explanations.

### 3.2.1 The Hypothesis Generation Process

We have structured the hypothesis generation process as a Tree of Thoughts (*35*), as illustrated in Fig. 2(b). A research paper is input, and an explanandum is subsequently extracted. The model formation process then abstracts the corresponding universals from that particular instance. Scientific laws are then associated with these universals, which are then evaluated and ranked to select the most plausible candidates. Finally, a scientific explanation is constructed based on the selected laws. The overall workflow is depicted in Fig. 3(b).

1) **Deductive-Nomological Decomposition:** Automated paper analysis and DN decomposition occur through LLM prompts. The standard structure of scientific research papers, i.e., the IMRAD format (*36*, *37*), provides a structured schema that is conducive to logical extraction. Here, the Introduction establishes the theoretical premises, the Methods define the antecedent conditions, the Results present the empirical phenomena, and the Discussion synthesizes the available explanations. This isomorphism allows LLMs to map passages of text to the DN model. Further validating this decomposability, recent frameworks like ResearchAgent (*19*) explicitly parse scientific contributions into problem (P), method (M), and experiment design (D), offering a semantic granularity that facilitates a mapping to the DN model's logical components.

2) **Modelling the Processes Involved in Generating Explananda:** Once the explanandum is extracted from a research paper, the next process should be modelling to extract the universals and have the LLM process them. However, as Salmon illustrates in (*11*), this process takes more time and space than events. This perspective fits with many of the concepts in basic formal ontology (BFO) (*38*, *39*). More precisely, BFO can serve as an upper-level ontological framework for reconstructing Salmon's distinctions between causal processes, causal interactions, and the transmission of conserved quantities. By Salmon's account (*12*), a causal interaction is an intersection of world-lines involving an exchange of a conserved quantity, while a causal process is the world-line of an object that transmits a non-zero amount of a conserved quantity through time. These notions align with BFO's categories of object, process, participant, and spatiotemporal region. Hence, any potential law that could explain the phenomenon can also be associated with the causal processes relating to the universals participating in the process. In fact, these instantiate the relevant properties or conserved quantities of the causal processes itself. Further, the universals in BFO are defined as a mind-independent, repeatable feature of reality that only exists when realized by one or more individual things (*38*). Therefore, each instance of those things can be mapped to the universals. Thus, the processes modelled by BFO effectively assemble the universals in preparation for associating any potential laws.

3) **Generating Associated Laws:** After collecting the universals in the process, the next step is to complete the ontology or knowledge graph using an LLM. The goal of this completion is to link the laws with their associated universals, that is, with the laws that apply to them. Following (*40*, *41*), we used context augmentation to identify the associated laws, where the contextual text retrieved from a vectorized data corpus of research papers is injected into the prompt to generate the laws.

4) **Law Evaluation and Ranking:** DN-Hypo-Pipeline uses LLMs-as-a-judge (*42*, *43*) to evaluate and score the associated laws according to a set of prompt templates. Hence, a subset of the top laws is selected, and an explanation is reconstructed for them. The scoring criteria include relevance, gap, logic, and feasibility, as outlined:

- Relevance: How relevant is the law to the phenomenon?
- Gap: Has the associated law already been proposed to account for this phenomenon?

- Logic: Does the law logically and coherently explain the phenomenon?
- Feasibility: Is it feasible to validate the proposed explanation?

More specifically, relevance assesses whether the law can shape the emergence of a phenomenon by applying its universal principles to any concrete entities or contexts that give rise to the phenomenon. Gap evaluates the degree of novelty and disruptiveness of the explanation. Logic examines the logical coherence and precision of the explanation. Lastly, feasibility gauges the practical difficulty of conducting validation experiments.

5) **(Re)constructing a Novel Law-based Explanation for the Explanandum:** The last step is to reconstruct the explanations according to the top-ranked associated laws. Given a recommended law as the core explanation, any other laws and conditions necessary for the explanation are constructed to form a deductive explanatory logic. To verify the novelty of the explanation, the framework incorporates retrieval augmented fact verification (*44*) implemented on a corpus of research papers.

# 4. Case study in the field of data science modelling

To test DN-Hypo-Pipeline, we applied the proposed framework to the field of data science modelling and evaluated the results in comparison to Baek et al.'s ResearchAgent (*19*) using both model-based evaluation and human assessment. Applying the pipeline to data modelling differs from other cases where scientific explanations might be developed because, typically, the purpose, i.e., the explanandum, of building a data model is to improve performance as opposed to finding some unknown underlying cause for a phenomenon. That said, the approach of using natural laws to elucidate and enhance model performance (or address related phenomena) is just as applicable and valuable to the realm of data science modeling as it is to other disciplines.

We must also note that a paradigm called "theory-guided data science" (*14*) has recently become popular in data science circles. This approach systematically integrates scientific theories into data science models during the process of knowledge discovery. However, although it is true that model building in data science often falls within the realm of mathematical modelling (*14*, *45*), the complete data science modelling process is a step beyond. Rather, the full process of data science modelling includes stages like data preparation, deployment, and communication, where mathematical modelling processes do not. However, mathematical modelling can support such stages, modelling "chosen" factors only "comes about by making some reasonable assumption based on a law of nature" (*46*). Moreover, as claimed by Ledder in (*47*), "a mathematical model is based on assumptions about the scientific principles that underlie the phenomena being modeled".

Importantly, we do not distinguish between "laws" and "principles" (*48*). Both are statements describing the nature of something and, in modelling, these are simply tools for making assumptions. Essentially, collecting or identifying the factors that influence a phenomenon is, in our proposed approach, merely a process of gathering the universals relevant to the situation. Therefore, our proposed framework is highly applicable to data modeling scenarios. Further, in essence, formulating assumptions is the link between the relevant laws and modelling, as illustrated in Fig. 4.

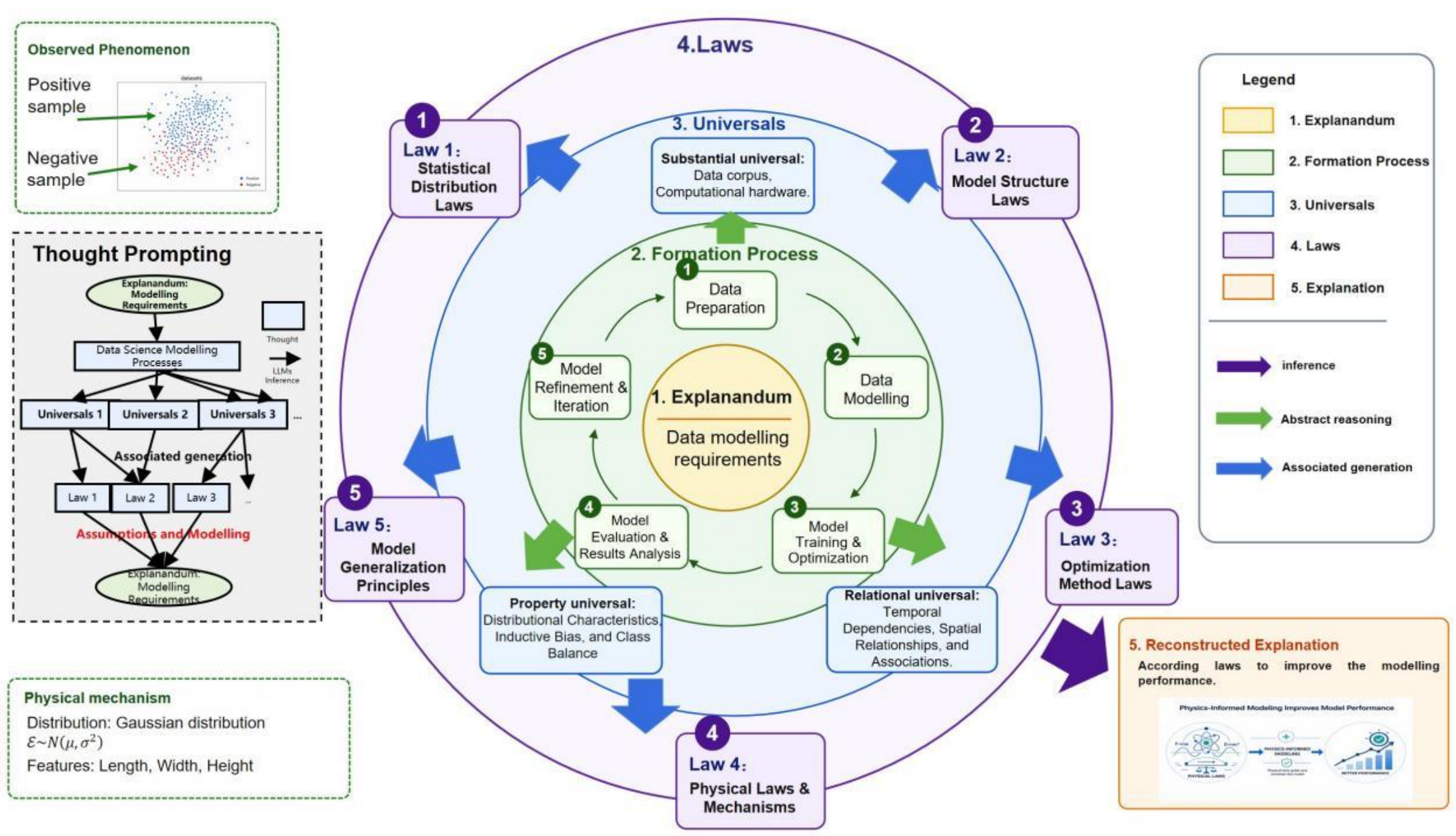

Fig. 4. Thought prompting in the data science model generation

## 4.1 Experimental Step

To test DN-Hypo-Pipeline, we selected the three most renowned papers in AI, generated a novel hypothesis to explain each problem, and devised an experimental plan to validate our approach. Further, we used four different LLMs to implement the experiments, comparing the results for each one. The four LLMs were GPT5.2, GROK4, DeepSeek3.2-Think, and Gemini-3.1-Pro. The selected papers describe three classical neural network structures: ResNet (*49*), Transformer (*50*), and Word Embedding (*51*). First, we observed the explanandum generation processes, i.e., the processes of data science modelling. We then analyzed the processes of generating the associated laws, followed by evaluating and ranking those laws. Last, we used RAG to check the novelty of the generated hypothesis based on the Arxiv dataset (*52*). Then, we selected the two highest-scoring hypotheses for experimental validation, which yielded a new model algorithm.

The selection of ResNet, Transformer, and Word Embedding is motivated primarily not by their influence, but by the fact that they exemplify three structurally distinct classes of explananda in data-science modeling—optimization and information-propagation pathways, sequential dependency modeling, and representation learning, respectively. As such, they constitute a differentiated test of our framework across phenomenon types, rather than a repetition of homogeneous samples. We position this evaluation as a proof-of-concept over a set of representative phenomenon families.

It is worth noting that all three are foundational, highly cited works whose extensive bodies of follow-up research have already explored and densely populated the surrounding space of possible innovations. This fact, far from making the task easier, raises the bar for proposing novel hypotheses: any hypothesis judged to be novel must clear a research frontier that has already been intensively saturated. Achieving results that significantly outperform direct generation on such highly competitive phenomena therefore constitutes a more stringent test of our framework, not a more lenient one. This is also consistent with our statistical findings: the significant main effect of the Paper factor on novelty and related dimensions indicates that the maturity of the research frontier surrounding a phenomenon systematically affects how difficult it is for a hypothesis to be rated as novel.

Additionally, to first validate that the pipeline ensures stable and convergent generation, we set the LLM temperature to 0.3, thereby reducing output diversity and uncertainty. Then, for benchmarking purposes against the direct generation capability of LLMs, we set the temperature to 0.7, thereby promoting higher innovativeness and diversity. Lastly, to ensure stability and reproducibility when LLMs serve as evaluators, we set the temperature to 0.1.

## 4.2 Evaluating the Generation of Universals

After Step 1) DN Decomposition, we modelled the explananda using BFO with the four LLMs. Here, we manually annotated the generated outputs to evaluate Precision, defined as the ratio of correctly generated ontologies validated against gold-standard criteria to the total number of ontologies produced by the model. We also calculated Relative Recall by aggregating the generated ontologies across all models. Relative Recall is the proportion of ontologies generated by a specific model relative to the total pool of ontologies collectively generated by all models. Both metrics were used to assess the quality and coverage of three processes involved in modeling the paradigms. The results are summarized in Fig. 5, following the methodology described in (*53*).

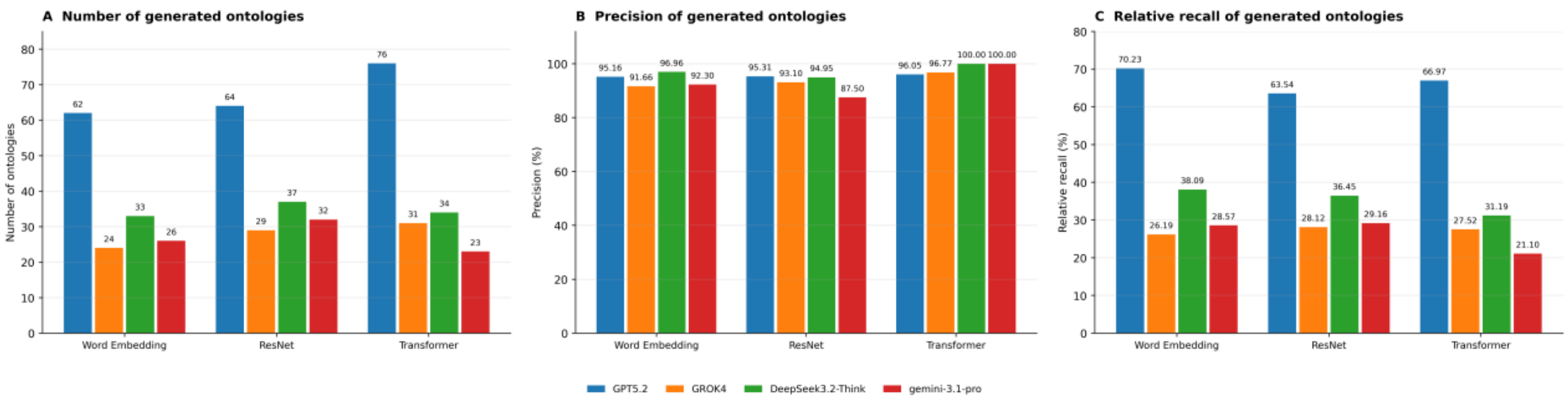

Fig. 5. Process-modeling performance across three target papers

From these results, we drew the conclusion that the observed Precision values are sufficiently high for process modeling tasks, indicating a strong ability to generate relevant ontologies. However, the lower relative recall values suggest that the individual LLMs struggled to comprehensively and precisely capture the intricate formation processes of the target phenomenon. This highlights their limitations in fully representing the underlying mechanisms of the explananda.

## 4.3 Evaluating the Generation of Laws

Next, we collected the universals and issued the prompts to generate the laws, evaluate them, and rank them using the four LLMs. The top five-ranked laws are presented in Fig. 6.

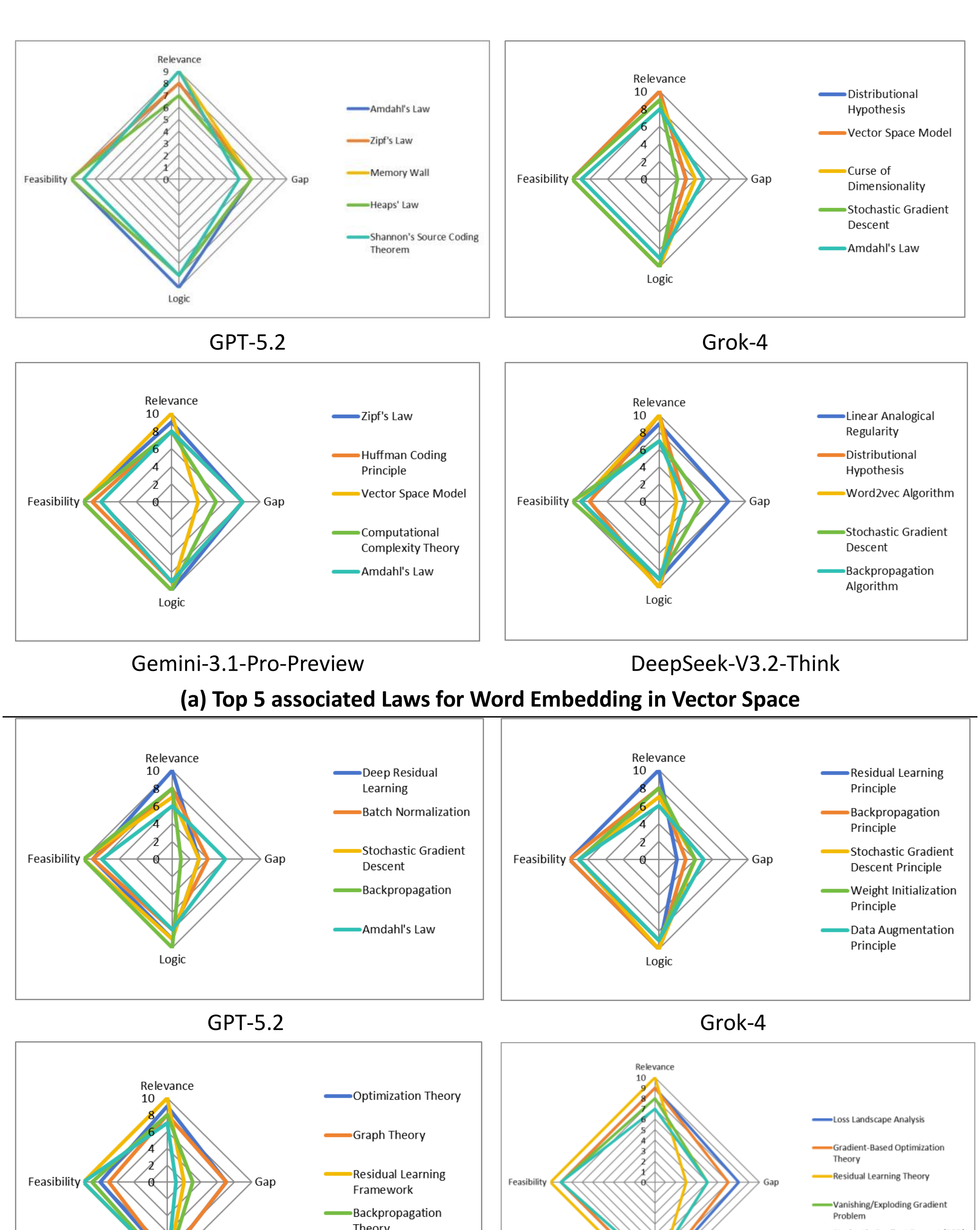

**(a) Top 5 associated Laws for Word Embedding in Vector Space**

Relevance
Feasibility
Gap
Logic
Deep Residual Learning
Batch Normalization
Stochastic Gradient Descent
Backpropagation
Amdahl's Law
GPT-5.2
Residual Learning Principle
Backpropagation Principle
Stochastic Gradient Descent Principle
Weight Initialization Principle
Data Augmentation Principle
Grok-4
Optimization Theory
Graph Theory
Residual Learning Framework
Backpropagation Theory
Chain Rule of Calculus
Gemini-3.1-Pro-Preview
Loss Landscape Analysis
Gradient-Based Optimization Theory
Residual Learning Theory
Vanishing/Exploding Gradient Problem
Stochastic Gradient Descent (SGD) Principle
DeepSeek-V3.2-Think

**(b) Top 5 associated Laws for ResNet**

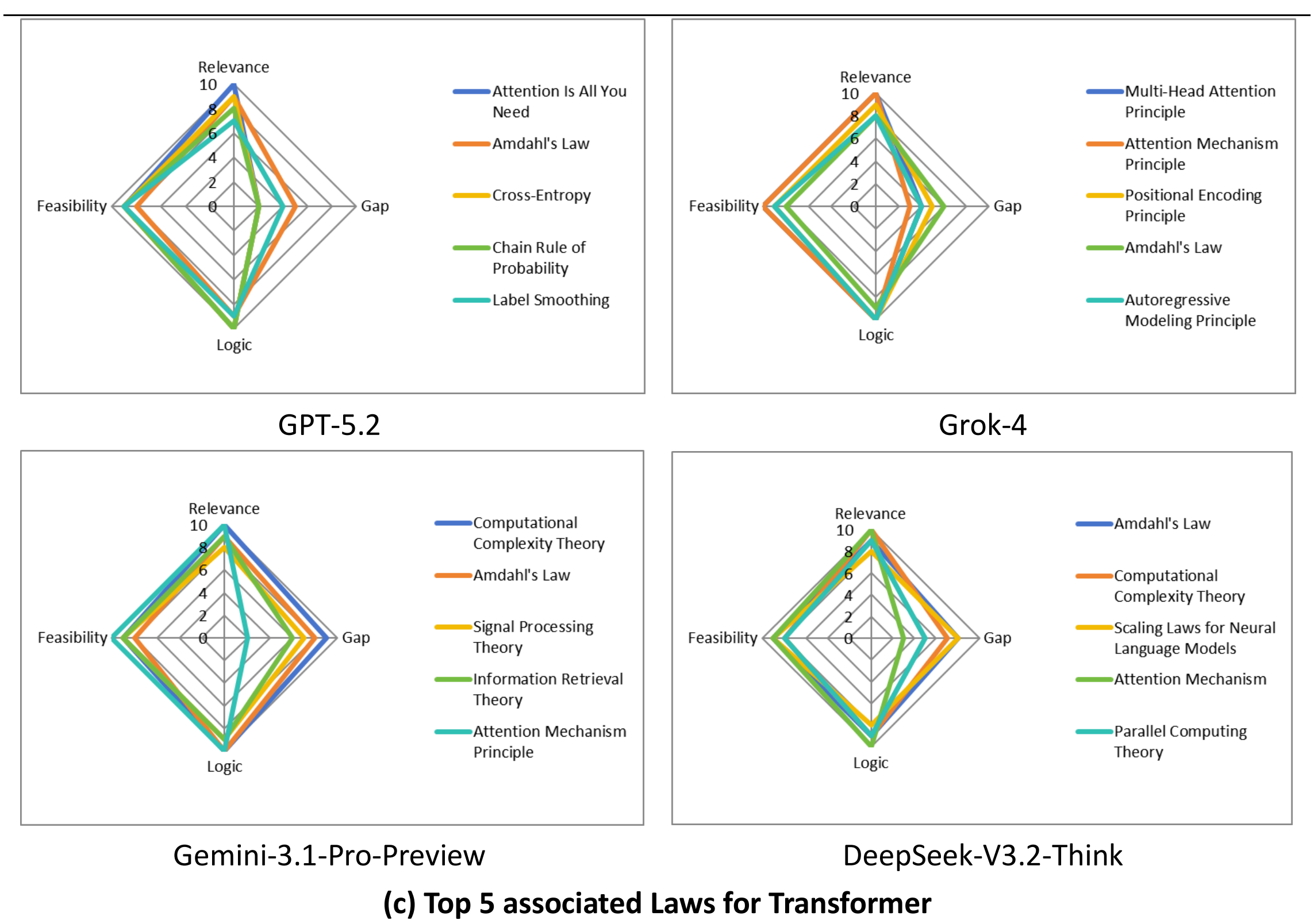

Fig. 6. Top 5 associated Laws for the three papers across four LLMs

Across the four LLMs tested on the word embedding paper (*51*), Amdahl's Law (*54*), Zipf's Law (*55*), and the distributional hypothesis consistently ranked the highest. Amdahl's Law pertains to the parallel execution of training workloads on GPU hardware, while Zipf's Law and the distributional hypothesis relate to the statistical properties of the training vocabulary within the corpus. For the ResNet paper, residual learning and backpropagation consistently ranked the highest. The low gap score reflects their long-standing maturity and extensive prior research, while their high composite score highlights their robust practical utility and empirical effectiveness. Similarly, with the Transformer paper, the attention mechanism exhibited a pattern identical to that of residual learning. While it consistently ranked highest, its low gap score reflects its conceptual maturity and extensive prior exploration, whereas its markedly high overall score highlights its proven practicality and foundational impact on modern architectures.

In summary, while variations in process modeling lead to model-specific theoretical associations, several classical and domain-relevant laws remained consistently prominent across all evaluated models.

## 4.4 Hypothesis Generation and Evaluation

The pipeline uses the identified theoretical laws to formulate and model a set of novel hypotheses. We performed this process with our three target papers using all four LLMs. To establish a comparative baseline, we also prompted the same models to generate hypotheses without any theoretical constraints. For evaluation, we devised a set of assessment criteria adapted from established research idea evaluation frameworks (*19*, *31*), as summarized:

- Validness: Rated from 1 to 5 based on theoretical grounding and feasibility.

- Novelty: Scored from 1 to 5 depending on the degree of originality and departure from existing work.
- Significance: Assessed from 1 to 5 according to the potential impact on the field.
- Potential: Judged from 1 to 5 for the breadth and promise of future research opportunities it presents.

The more detailed scoring rubrics we developed for each metric can be found in the Appendix E.

Specifically, we selected the top five ranked laws to use as a basis for generating the explanations, with each law yielding one distinct proposal. Each of the four LLMs independently generated ideas for all five laws across the three papers, resulting in 5 × 4 × 3 = 60 law-guided modeling ideas. Additionally, for each paper, the four LLMs were prompted to directly generate one baseline idea each, yielding 1 × 4 × 3 = 12 direct-generation ideas. Therefore, in total, 72 modeling ideas were produced for evaluation.

We followed the LLM-as-a-judge paradigm (*42*, *43*), automating evaluation using four distinct LLMs, complemented by a comparison group of three human experts (All PhD holders in computer science or related fields specializing in AI). Consequently, each proposal was independently evaluated seven times. To ensure robust assessment and facilitate direct comparison, all instances were scored independently by these seven evaluators. To control for potential evaluation bias caused by differences in the level of detail between the two generation methods, we introduced an ablation evaluation setting. In this setting, the four judge LLMs were explicitly instructed to evaluate only the core conceptual idea of each output, while disregarding any recommended implementation details. This design allows us to isolate the intrinsic quality of the ideas from the confounding effect of differences in elaboration depth. The scores from the LLMs and experts were summed separately to yield distinct final metrics for each group. Further, to enhance the interpretability and reliability of the scoring process, both the judge-LLMs and the human experts were asked to provide structured rationales to accompany their assigned scores.

## 4.4.1 Direct VS. DN-Hypo-Pipeline Generation

We formulated three null hypotheses to statistically compare the efficacy of the law-guided hypotheses to the direct baseline. These three hypotheses follow:

- $H_0^{(1)}$: The maximum score among the five law-guided modeling proposals is less than or equal to the score of the directly generated baseline idea.
- $H_0^{(2)}$: The mean score of the five law-guided modeling proposals is less than or equal to the score of the directly generated baseline idea.
- $H_0^{(3)}$: The median score of the five law-guided modeling proposals is less than or equal to the score of the directly generated baseline idea.

The statistical results of the Wilcoxon signed-rank test (*56*) are shown in Table 4. At the $\alpha=0.05$ significance level, $H_0^{(1)}$ was rejected by both the LLMs and humans because the maximum score among the law-guided proposals was statistically significantly higher than that of the directly-generated baseline idea. $H_0^{(2)}$ was also rejected by both evaluator groups since the mean evaluation score of the law-guided proposals was statistically significantly higher than that of the directly-generated baseline. In terms of $H_0^{(3)}$, the analysis yielded a non-significant result ($p=0.0859$) by the LLMs, providing insufficient evidence to assert a median of >0 under the 0.05 threshold. By contrast, the humans produced a statistically significant result, leading to the opposite conclusion. Thus, both LLM-as-Judge and Human Expert as judge prove the DN-Hypo-Pipeline is better than direct generation. In Fig. 7(a) diff_mean is the mean of the differences between the two statistics.

To further verify that this advantage stems from the intrinsic quality of the core concepts rather than from differences in the level of elaboration, we introduced an ablation setting termed "Idea-Only LLM-as-a-Judge". In this setting, the judge LLMs were explicitly instructed to evaluate only the core conceptual idea of each proposal, while disregarding any recommended implementation details. The statistical results of this ablation, shown in Fig. 7(a), indicate that even when implementation-specific details are excluded from consideration, the law-guided proposals still achieve significantly higher conceptual scores than the direct baselines. This finding confirms that the proposed approach improves research ideation at the conceptual level, independent of the structural depth or implementation detail of the generated text.

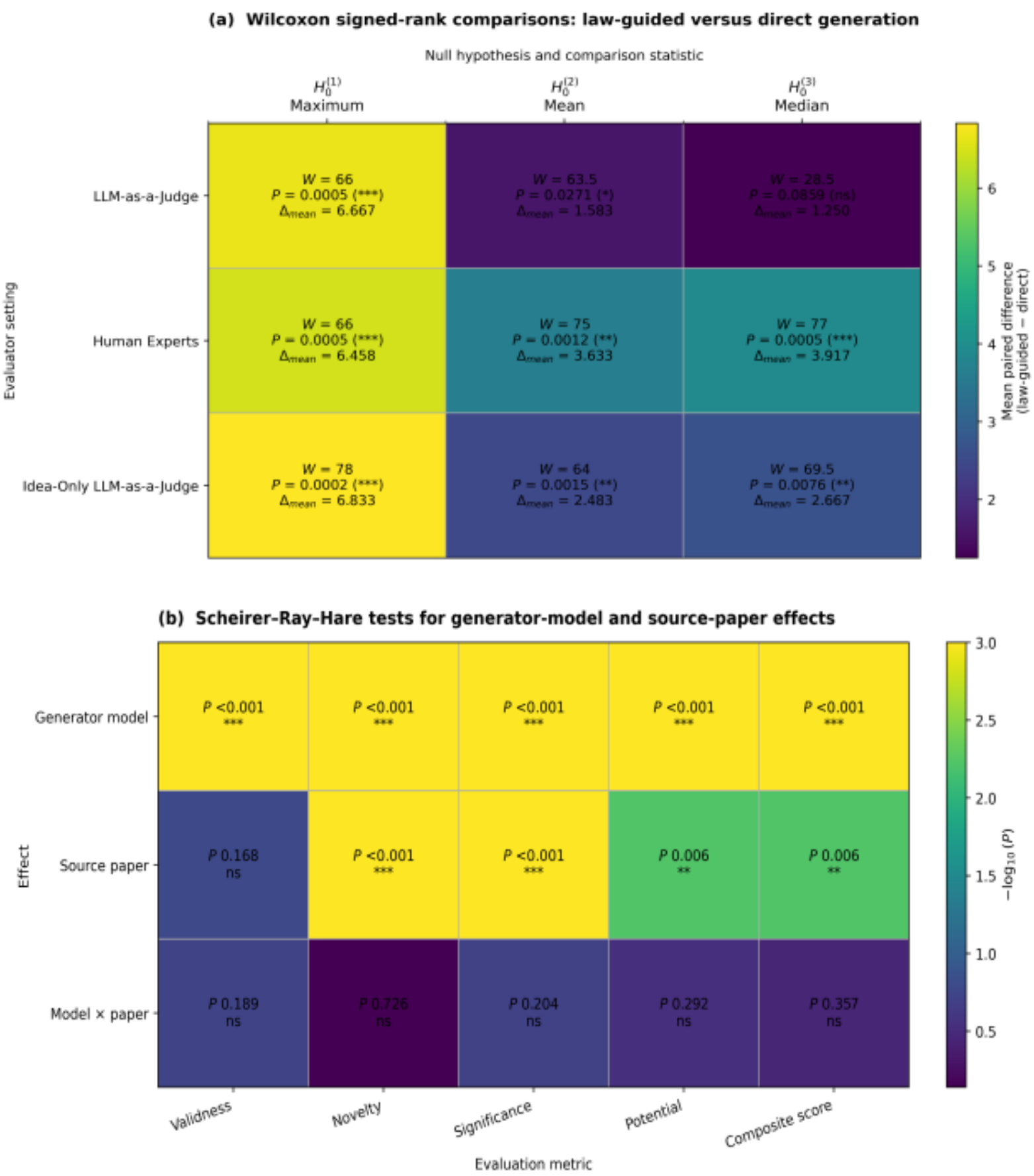

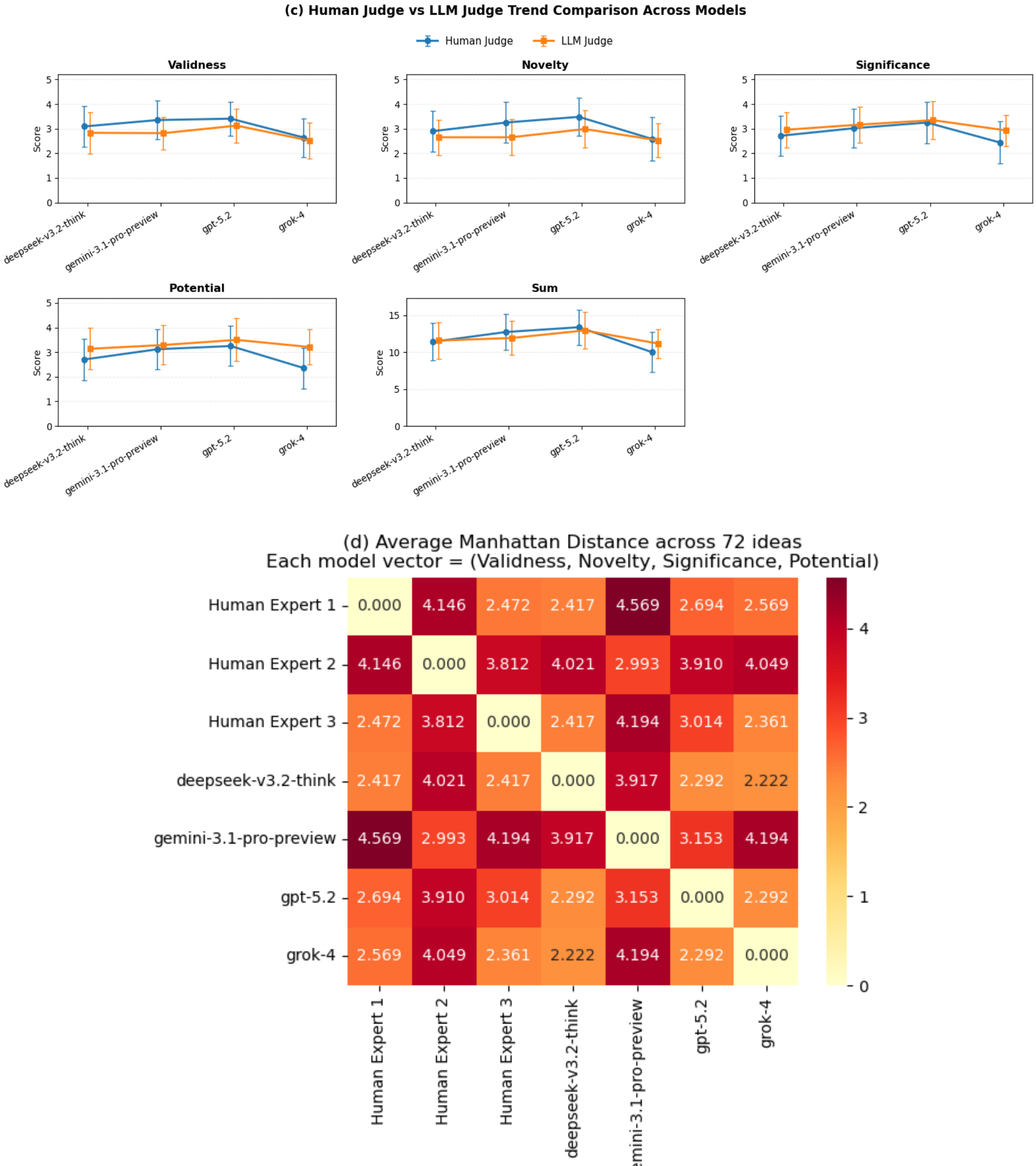

Fig. 7. Statistical evaluation of hypothesis-generation performance and evaluator agreement. (a) Wilcoxon signed-rank test results comparing law-guided hypothesis generation with direct generation under three criteria: the maximum, mean, and median scores of the five law-guided hypotheses. (b) Scheirer–Ray–Hare test results for the effects of the hypothesis-generating model, source paper, and their interaction on validness, novelty, significance, potential, and composite score. (c) Mean scores and standard deviations across evaluation metrics and hypothesis-generating models, as assessed by LLM judges and human experts. (d) Average pairwise Manhattan distances across 72 generated ideas among four LLM judges and three human experts.

### 4.4.2 Comparative Evaluation of Hypothesis Generation

We further compared the generative capabilities of the four evaluated LLMs, with the corresponding descriptive statistics, as summarized in Fig. 7(c). Across all assessed metrics, GPT-5.2 consistently achieved the highest performance, followed by Gemini-3.1-Pro-Preview as the second-best model. Observing that the evaluations from the LLMs and human experts generally led to the same conclusion demonstrates the reliability of using LLMs as evaluators.

To assess the statistical main and interaction effects of the generative models (i.e., the four LLMs serving as hypothesis generators, as opposed to judges) and the prior-start factor (i.e., the three input papers), we performed Scheirer-Ray-Hare tests (*57*) on a total of 504 individual scores. Note the total of 504 was derived as follows: 35 scores per 12 distinct evaluation instances derived from 5 generated law-based hypotheses × [4 LLM-as-judge + 3 human experts], plus 7 scores per 12 distinct evaluation instances for the directly-generated baseline hypotheses, derived from 1 generated hypothesis × the same 7 evaluators. The tests were applied to each of the four metrics reported at the beginning of section 4.4 as well as to their composite score. The Scheirer-Ray-Hare test (*57*) was chosen as a non-parametric alternative to a two-way ANOVA because the scores did not follow a normal distribution and the design involved unequal cell sizes (35 vs. 7); although they remained robust under heteroscedasticity and ordinal-scale measurements.

As reported in Fig. 7(b), the LLM factor exhibits a significant main effect across all metrics, confirming inherent differences in the models' idea-generation capabilities. The paper factor significantly influences only the novelty, significance, and potential metrics, suggesting that certain research trajectories have reached conceptual saturation. Notably, the absence of a significant LLM × Paper interaction indicates that the relative performance of the LLMs remains stable across different papers. In turn, this implies that any differences in generative capacity are intrinsic rather than context-dependent.

### 4.4.3 Comparison Between the LLMs and Humans as Judges

To quantify inter-evaluator consistency between the four LLM judges and the three human experts, we calculated the pairwise Manhattan distances (*58*) across their multidimensional scoring profiles for each generated idea. The average distance is reported as a measure of scoring divergence. This metric is particularly well-suited to capturing absolute discrepancies across multiple evaluation criteria. As illustrated in Fig. 7(d), the human expert 1, human expert 3, GPT-5.2, Grok-4, and DeepSeek-V3.2-Think show consistently low pairwise distances. This indicates strong alignment in their assessment frameworks. By contrast, Gemini-3.1-Pro-Preview and human expert 2 form a tight cluster with low inter-distance, but demonstrates substantially higher distances relative to the other evaluators, reflecting a distinctly different evaluation paradigm.

## 4.5 Selected Conceptual Idea Validation

We selected the two concept proposals with the highest scores to validate our proposed method. The first was to improve the Transformer's computation complexity. The second was to improve the word embedding algorithms Skip-Gram and CBOW (*51*).

### 4.5.1 Continuous-Time Attention Transformer

We developed a solution with a high evaluation score called CTAT (Continuous-Time Attention Transformer), generated from DeepSeek. Inspired by the concept of attention, the solution incorporates continuous-time attention in the form of cumulative kernel functions. These speed up the attention computations via numerical integration while also reducing the time complexity as follows:

$$\mathrm{y}(\tau) = \int Kernel(\tau, t) v(t) dt \xrightarrow{\text{instead}} Attention(Q, K, V) = softmax(\frac{QK^T}{\sqrt{d_k}})V \quad (5)$$

$$Kernel(\tau, t) = \frac{\exp\left(\frac{q(\tau)^T k(t)}{\sqrt{d_k}}\right)}{\int \exp\left(\frac{q(\tau)^T k(s)}{\sqrt{d_k}}\right) ds} \quad (6)$$

where $Q/q(t), K/k(t), V/v(t)$ represent the query, key, and value in the attention mechanism (*50*). Note that discrete forms of Eqs. (5) and (6) were published and analyzed in (*59*). So, here, when each sample point corresponds to an individual token, the implementation becomes equivalent to that in (*59*). However, CTAT uses a Gaussian kernel as the integration kernel for position encoding, which can be interpreted as a temporal convolution kernel. Efficient numerical integration schemes for Gaussian kernel functions include Gauss–Legendre quadrature (*60*) and the Fast Fourier Transform (FFT) (*61*). However, the Fast Fourier Transform requires a linear kernel function (*62*) to replace the exponential function in the continuous-time softmax attention mechanism defined in (5) and (6). Therefore, we devised a softmax attention mechanism where the Gaussian kernel is evaluated via the Gauss–Legendre quadrature, along with a linear attention variant in which the Gaussian kernel is evaluated by both the Gauss–Legendre quadrature and the Fast Fourier Transform.

Following Tsai et al. (*59*), we evaluated the proposed model exclusively on a sequence prediction task using a decoder-only architecture with the WikiText-2 dataset. This meant we could rapidly assess its regression capability and also draw a direct comparison between the actual training time and the theoretical time complexity.

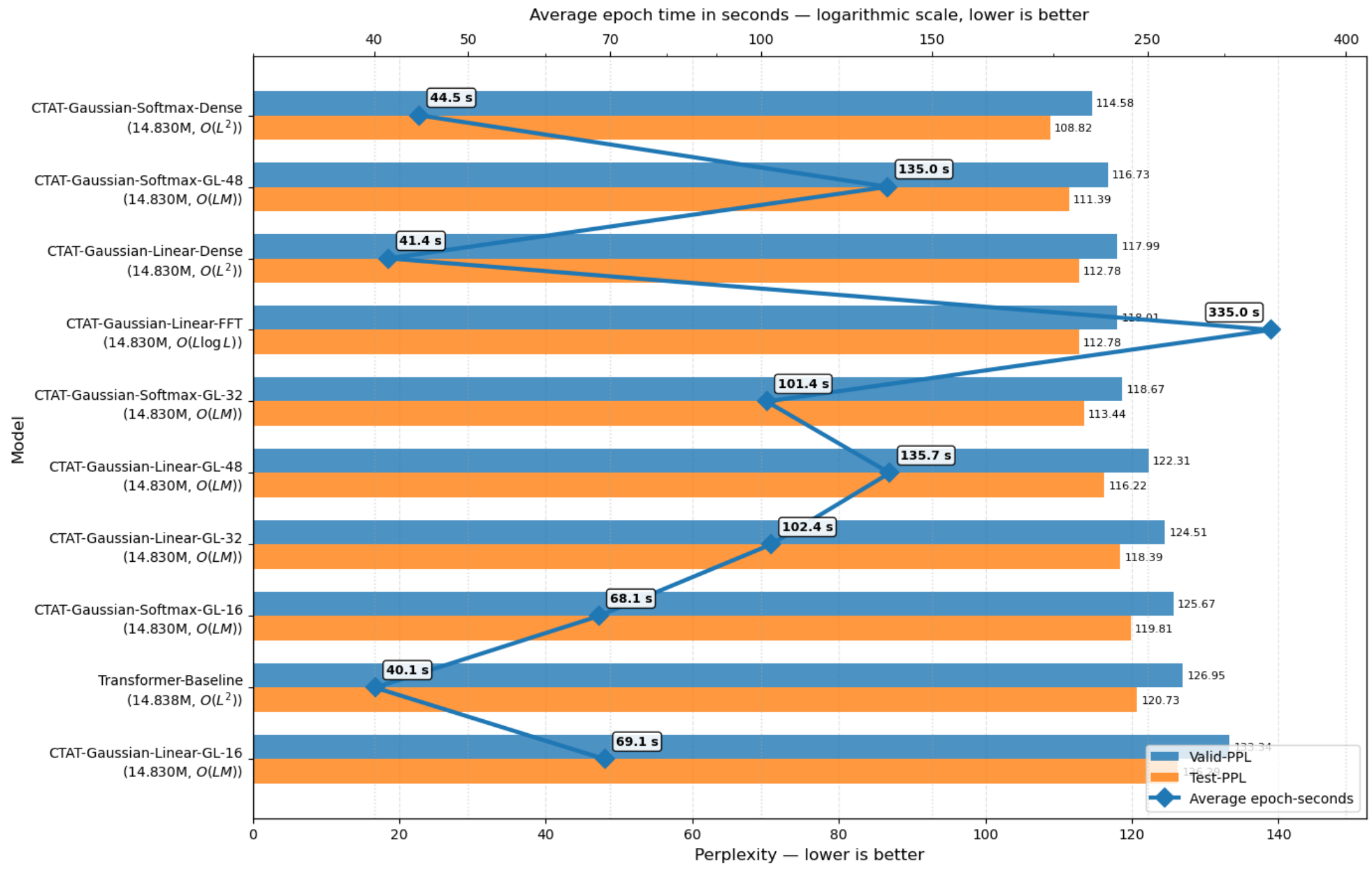

Bars represent Valid/Test PPL. Diamond markers represent average epoch time. Model labels include number of parameters and theoretical complexity.

Fig. 8. Results of the sequence prediction task with the WikiText-2 dataset

The experimental results are summarized in Fig. 8, where PPL denotes perplexity averaged over five random seeds. The best-performing configuration was CTAT-Gaussian-Softmax-Dense, which involves a Gaussian kernel with standard softmax attention and per-token sampling. This closely resembles the discrete-time Gaussian positional encoding proposed in (*59*). CTAT-Gaussian-Linear-Dense is its linear-attention counterpart. Both configurations achieve a lower test PPL than the Transformer baseline.

In terms of the numerical integration, the Gauss-Legendre collocation method (CTAT-Gaussian-Softmax-GL-48) delivered the best performance with $M = 48$, where $M$ denotes the number of interpolation nodes. The Fast Fourier Transform method (CTAT-Gaussian-Linear-FFT) achieves accuracy comparable to that of the original dense sampling scheme for linear attention functions.

In terms of computational complexity, the Transformer baseline, implemented via PyTorch's native Transformer functions and thoroughly optimized, yielded the fastest training throughput. Fast Fourier Transform was the slowest method in practice; however, its theoretical time complexity is $O(L \log L)$, where $L$ denotes the sequence length. Hence, for sequences in the order of $10^6$ tokens, $\log L < 48$, rendering with this method would theoretically be no worse than the Gauss-Legendre method. Plus, Fast Fourier Transform has the greatest potential for further optimization among all methods considered. The Gauss-Legendre method is equally promising, as its complexity scales linearly with $L$. Also, it only incurs a negligible accuracy loss (108.82 to 111.39) compared to dense sampling.

CTAT is similar to the continuous-time transformer ContiFormer in (*63*), but it has some important differences. First, due to its continuous domain, CTAT is a “kernel method generalization” of the discrete Transformer self-attention mechanism. Second, the attention score is a kernel function, and the final attention computation is an integral aggregation (i.e., a continuous integration over a source time) with continuous weights. By contrast, ContiFormer is an adaptation of the Transformer architecture to continuous signals with irregularly sampled time points. It assumes that the keys/values are signals evolving over time, which can be modeled by ODEs as continuous evolution functions. Further, the final attention computation is a discrete aggregation (i.e., a summation over discrete source indices) with continuous weights.

### 4.5.2 Heaps-Adaptive Lexico Manifold Embedding

Next, we implement a word embedding solution called HALO (Heaps-Adaptive Lexico Manifold Embedding) generated by GPT-5.2. This solution organizes a vocabulary into high-frequency core words and low-frequency tail words based on Heaps' law, which ranks words according to their frequency. Core words retain independent, explicit word vectors, while tail words forgo a separate embedding. Instead, their vectors are dynamically generated using a shared generator based on character n-grams, which multiplies the parameter compression for long-tail words. Once low-frequency words are allocated to the tail set, they no longer correlate to individual word vectors. Rather, their representations are dynamically synthesized through a shared generator that follows the process of character n-gram processing and subword pooling. They become coefficient-based and are combined with residuals to result in significantly fewer parameters.

Following Mikolov et al. (*51*), we implemented HALO with both the Continuous Bag-of-Words (CBOW) and Skip-Gram with Negative Sampling (SGNS) training modes, training all models on the WikiText-103 training set (*64*). To evaluate this solution, we assigned the models three tasks: the Google analogy task (*65*), the Stanford RareWord Similarity task (*66*), and the WordSim-353 task (*67*). We ran the baseline CBOW and SGNS models with three random seeds and averaged their results. With the HALO variants, we evaluated all the models in a single run. Each variant was trained for 40 epochs, while the baseline models were trained until they converged. The reported performance values were taken from the epoch with the highest accuracy on the Google analogy task. Also, we only varied the dimensionality of the word representations – all other hyperparameters, such as the size of the context window and the number of negative samples, were fixed.

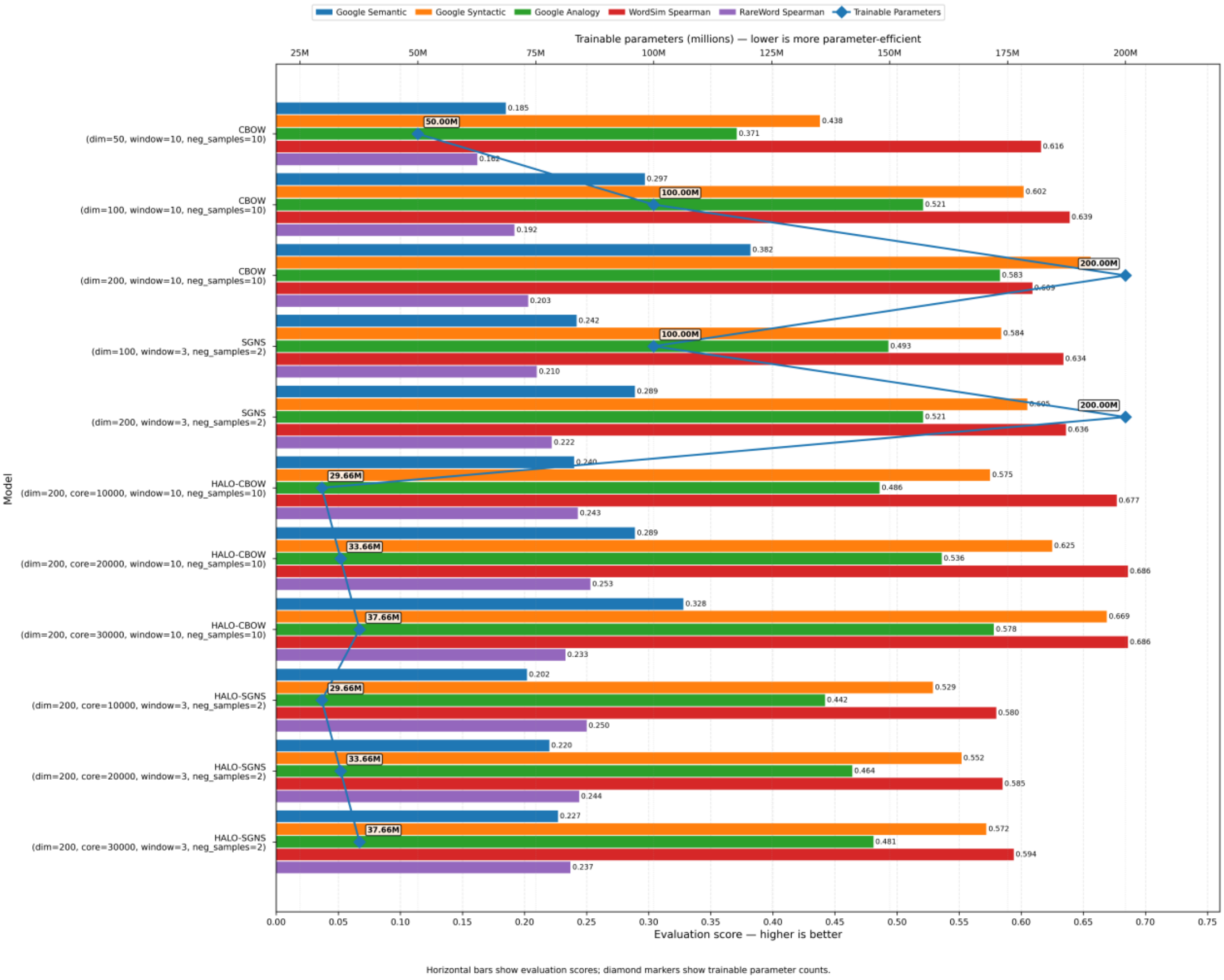

Fig. 9. Word embedding experiment results with the WikiText-103 dataset

The experimental results are summarized in Fig. 9. The HALO-CBOW model (with an embedding dimension of 200 and a core vocabulary size of 30,000) performed the best, which is particularly notable given that it used so few parameters. As the core vocabulary size increased, HALO-CBOW's performance improved with the Google analogy and word similarity tasks. However, it decreased with the rare word similarity task as the core size grew larger. This is because expanding the core vocabulary reduces the number of low-frequency tail words, which are the main contributors to rare word similarity performance. Another interesting observation we made from the results was that HALO-SGNS did not converge well, nor did it perform well overall.

HALO is closely related to rare word modeling approaches that are based on subwords, as well as embedding compression methods like fastText (*68*), MIMICK (*66*), hash embeddings (*69*), adaptive input representations (*64*), and compositional code learning (*67*). However, unlike fastText, HALO does not simply sum n-gram vectors; it generates tail-word embeddings through a subword-conditioned manifold generator. Further, unlike adaptive input representations, HALO models rare words with an implicit generator rather than lower-dimensional frequency clusters.

# 5. Experimental Results Analysis and Discussion

The experiments show that not only can LLMs propose hypotheses, but also the best hypotheses generated by DN-Hypo-Pipeline improved the aggregate sum of scores by an average of more than 6 points (53-46.33=6.67) across a total of 80 scores (4 LLMs * total 20 scores), as outlined in Fig. 7(a). Additionally, Figs. 7(c) show that human experts and LLMs both yield largely consistent overall rankings, with GPT5.2 emerging as the best-performing model. In terms of the evaluations, three of the four scores assigned by the panel of LLM judges were consistent with those given by the two of three human experts. With the exception of Gemini, small Manhattan distances indicated high agreement. This validates the reliability of the LLM-as-a-judge approach.

Additionally, all the models generated similar core ontological processes, but the relevant ontologies generated along with their associated theories, laws, and principles varied significantly. Nevertheless, the high-scoring laws had some convergence, which in turn yielded high-scoring solutions. However, no quantitative metric currently exists to evaluate this dimension.

In terms of validation, we selected the two highest-scoring solutions and compared them with the proposals in the original papers. CTAT significantly reduced the theoretical time complexity with only a marginal drop in accuracy. Although the actual runtime was slower than the baseline due to a lack of low-level operator optimizations, the results do demonstrate CTAT as a promising direction for future operator-level optimizations. HALO, with the Google analogy task, maintained accuracy and achieved superior results with the rare-word task with significantly fewer parameters.

To further validate DN-Hypo-Pipeline, we manually checked the literature to determine whether the two top-scoring hypotheses generated were truly novel. As such, we searched Google Scholar using combinations of the hypotheses' core concepts and keywords, then manually reviewed the titles and abstracts on the first page of the search results one by one to determine whether any highly similar or substantially equivalent prior works existed. None did.

Additionally, as a contribution to the field of data modelling, DN-Hypo-Pipeline extends the principles of theory-guided data science (*14*) and physics-informed machine learning (*70*). As highlighted by Karniadakis (*70*), the core insight of theory-guided and physics-informed Paradigm lies in embedding the physical principles and the domain laws into a model's design to ensure physical consistency. This embedding is accomplished by designing network architectures, integrating penalty constraints, and/or applying data augmentation techniques, which collectively introduce observational biases, inductive biases, and learning biases that encompass the entire data modeling process. Our methodology not only spans the complete workflow of data modeling, but it also systematically handles biases in Step 2) Modelling the Processes Involved in Generating Explananda. Further, Step 3) Generating the Associated Laws involves collecting any laws or principles relevant to the generated explanandum and then devising an appropriate strategy to embed laws in modelling the explanandum in Step 5) (Re)constructing a Novel Law-based Explanation for the Explanandum. For example, in the case of the word embedding explanation in Figs. 7 (b) and (c), Zipf's Law describes single-word distributions, while the Distributional Hypothesis (*51*) describes the word co-occurrence distributions. However, integrating these theories into the models' assumptions could introduce inductive biases, whereas our method for collecting laws and principles is well-founded theoretically. Moreover, it exhibits completeness. As Salmon (*11*) argues, the laws associated with the model are in the process of formulating the explanandum (the process of data science modelling). Further, as proposed by Karpatne et al. (*14*), theory-guided approaches are superior to purely data-driven approaches based on the following principle: physical consistency serves as a critical component of model performance, alongside training accuracy and model complexity.

$$\text{Performance} \propto \text{Accuracy} + \text{Simplicity} + \text{Consistency}.$$

Although our proposed approach is based on theoretical deduction, it can be recognized as a generalization of the theory-guided / physics-informed modelling paradigm to other natural science domains. Thanks to the DN-model (*34*), theory can explain a phenomenon in addition to being used for modeling. In other words, modelling a phenomenon is also an explanation for the phenomenon.

# 6. Limitations and Conclusions

The limitations of our approach largely parallel the limitations of LLMs. When using an LLM to generate open-ended ontologies, such as when generating universals and laws, hallucinations can manifest in a particularly stubborn and intractable form. Hence, when a model is required to construct a complete conceptual system from scratch, including entities, attributes, relations, hierarchies, and axioms, based on the objectives of modelling the processes associated with formulating an explanandum, the lack of an external gold standard transforms hallucination from a mere factual fabrication into the systematic generation of a conceptually “profound and self-consistent” framework. Compounding this issue, the number of entities and relations generated varies dramatically across models, and, beyond hallucination, there is not even a gold standard for completeness. That is, there is no agreed-upon benchmark for what constitutes a "sufficiently comprehensive" ontology.

Another limitation concerns the limitations of the DN model. Distinguishing genuine laws of nature from accidentally true generalizations remains a classic, unresolved dilemma within the DN model. Indeed, Hempel himself conceded that characterizing “lawlikeness” is a “highly recalcitrant” problem (*33*). In our pipeline, laws are proposed to generate based on the scientific literature, which contains a vast array of valid scientific facts. Unfortunately, LLMs mostly fail to surmount this epistemic hurdle. They struggle to reliably differentiate generated rules with genuine nomic status from those that are merely accidental generalizations of scientific facts extracted from the literature. Consequently, while the generated “laws” may formally comply with the requirements of the DN model, they substantively amount to nothing more than sophisticated, but spurious laws.

A third limitation is methodological and concerns the layered explanation-theoretic design itself. Our integration of the three accounts is, by construction, a division of labor adopted for engineering feasibility—DN fixing the output form, Salmon constraining the search space, and Armstrong bridging universals to laws—rather than a demonstrated reconciliation of them at the level of the philosophy of explanation. Two consequences follow. First, Salmon's requirement that the admitted universals correspond to causal processes actually present in the phenomenon's formation is, in our pipeline, adjudicated by the LLM's own ontological modeling under BFO; in the absence of an external gold standard for what processes truly exist, this constraint degenerates in practice into a criterion of LLM self-consistency rather than a strict ontological guarantee. Second, the notions of causal process and conserved-quantity transmission were originally articulated for physical systems, and their application to data-science phenomena (e.g., the statistical regularities captured by Zipf's or Heaps' law) is at present analogical rather than strict, and we have not separately established their adequacy in this non-physical domain. We therefore regard DN-Hypo-Pipeline as an initial, working operationalization that demonstrates feasibility and yields verifiable algorithmic outcomes, while leaving the deeper questions of philosophical coherence and cross-domain justification to future work.

Regarding the scope of evaluation, our study validates the framework on three explananda. While these cover structurally distinct phenomenon types, as discussed in Section 4.1, their limited number constrains the external validity of cross-paper statistical inferences—such as the Paper-factor effect in the Scheirer-Ray-Hare test. Broader coverage across a larger set of phenomena is left to future work.

In this paper, we propose a scientific hypothesis generation pipeline based on the DN model that has been practically implemented and validated within the data science domain. From an analysis of three classic highly-cited papers, we derived two novel solutions: one to reduce the computational complexity of Transformer models, and another to vectorize words based on frequency. The contributions of this work can therefore be summarized as follows:

- First, we developed a scientific hypothesis generation pipeline called DN-Hypo-Pipeline, which is based on the DN model. This pipeline is derived from scientific principles as opposed to empirical or heuristic methods, thus representing a principled and convergent approach to hypothesis generation.
- Second, we demonstrated that the hypotheses generated by the pipeline outperformed those produced by direct generation. We also compared the performance of various LLMs in pipeline generation, identifying GPT 5.2 as the best model. Additionally, we compared human hypothesis evaluation with LLM-as-a-judge evaluation. With the exception of Gemini-3.1-pro-preview, the LLMs' judgements were consistent with the human experts as measured by Manhattan distances.
- Third, we validated two of our generated hypotheses by developing two novel algorithms: one designed to accelerate Transformer computation with minimal loss of accuracy, and the second to vectorize words using a separate expression space based on frequency.

# 7. References

1. scientific method. https://en.wikipedia.org/wiki/Scientific_method.
2. scientific-method. https://plato.stanford.edu/entries/scientific-method/.
3. C. G. Hempel, *Philosophy of Natural Science* (Prentice Hall, 1966).
4. hypothesis. https://en.wikipedia.org/wiki/Hypothesis
5. S. Ren, P. Jian, Z. Ren, C. Leng, C. Xie, J. Zhang, Towards Scientific Intelligence: A Survey of LLM-based Scientific Agents. (2025).
6. J. Gottweis, W.-H. Weng, A. Daryin, T. Tu, P. Sirkovic, A. Myaskovsky, G. Glowaty, F. Weissenberger, A. Orlandi, D. Popovici, A. Palepu, K. Rong, R. Tanno, K. Saab, F. Zhang, J. Blum, A. Carroll, K. Kulkarni, N. Tomašev, D. Zverinski, I. Rendulic, E. Vedadi, F. Hasler, L. Rimanic, M.

Boia, I. Budiselic, B. Feinstein, M. Bellaiche, T. Sheffer, J. Freyberg, J. Ratcliff, O. Bertolli, K. Chou, A. Hassidim, B. Gokturk, A. Vahdat, Y. Guan, V. Dhillon, E. D. Vaishnav, B. Lee, T. R. D. Costa, J. R. Penadés, G. Peltz, Y. Matias, J. Manyika, D. Hassabis, Y. Xu, P. Kohli, A. Pawlosky, A. Karthikesalingam, V. Natarajan, Accelerating scientific discovery with Co-Scientist. *Nature*, doi: 10.1038/s41586-026-10644-y (2026).

7. C. Lu, C. Lu, R. T. Lange, Y. Yamada, S. Hu, J. Foerster, D. Ha, J. Clune, Towards end-to-end automation of AI research. *Nature* **651**, 914–919 (2026).
8. Z. Wang, B. Danek, Z. Yang, Z. Chen, J. Sun, Can Large Language Models Replace Data Scientists in Clinical Research? *Arxiv*, 1–28 (2024).
9. H. W. Sprueill, C. Edwards, K. Agarwal, M. V Olarte, U. Sanyal, C. Johnston, H. Liu, H. Ji, S. Choudhury, "CHEMREASONER: heuristic search over a large language model's knowledge space using quantum-chemical feedback" in *Proceedings of the 41st International Conference on Machine Learning* (JMLR.org, 2024)*ICML'24*.
10. C. Cao, X. Cao, M. Cashman, M. Kumar, A. Timoshenko, J. Yang, S. Yu, J. Zhang, Y. Zhu, B. Wernerfelt, How do successful scholars get their best research ideas? An exploration. *Mark. Lett.* **30**, 221–232 (2019).
11. W. C. Salmon, W. C. Salmon, *Scientific Explanation and the Causal Structure of the World* (Princeton University Press, Princeton, 2020; https://doi.org/10.1515/9780691221489).
12. W. C. Salmon, Causality and Explanation: A Reply to Two Critiques. *Philos. Sci.* **64**, 461–477 (1997).
13. D. M. ARMSTRONG, Laws of Nature As Relations Between Universals, and As Universals. *Philos. Top.* **13**, 7–24 (1982).
14. A. Karpatne, G. Atluri, J. H. Faghmous, M. Steinbach, A. Banerjee, A. Ganguly, S. Shekhar, N. Samatova, V. Kumar, Theory-Guided Data Science: A New Paradigm for Scientific Discovery from Data. *IEEE Trans. Knowl. Data Eng.* **29**, 2318–2331 (2017).
15. I. Ciucă, Y.-S. Ting, S. Kruk, K. Iyer, Harnessing the Power of Adversarial Prompting and Large Language Models for Robust Hypothesis Generation in Astronomy. (2023).
16. T. O'Brien, J. Stremmel, L. Pio-Lopez, P. McMillen, C. Rasmussen-Ivey, M. Levin, Machine learning for hypothesis generation in biology and medicine: exploring the latent space of neuroscience and developmental bioelectricity. *Digit. Discov.* **3**, 249–263 (2024).
17. B. Qi, K. Zhang, K. Tian, H. Li, Z.-R. Chen, S. Zeng, E. Hua, H. Jinfang, B. Zhou, Large Language Models as Biomedical Hypothesis Generators: A Comprehensive Evaluation. (2024).
18. M. Radensky, S. Shahid, R. Fok, P. Siangliulue, T. Hope, D. S. Weld, "Scideator: Human-LLM Compound System for Scientific Ideation through Facet Recombination and Novelty Evaluation" in *Proceedings of the ACM Conference on AI and Agentic Systems* (Association for Computing Machinery, New York, NY, USA, 2026; https://doi.org/10.1145/3786335.3813161), pp. 348–374.
19. J. Baek, S. K. Jauhar, S. Cucerzan, S. J. Hwang, ResearchAgent: Iterative Research Idea Generation over Scientific Literature with Large Language Models. *Proc. 2025 Annu. Conf. Nations Am. Chapter Assoc. Comput. Linguist. Hum. Lang. Technol. Long Pap. NAACL-HLT 2025* **1**, 6709–6738 (2025).
20. C. O'Neill, T. Ghosal, R. Răileanu, M. Walmsley, T. Bui, K. Schawinski, I. Ciucă, Sparks of Science: Hypothesis Generation Using Structured Paper Data. (2025).
21. R. Li, L. Jing, C. Han, J. Zhou, X. Du, Learning to Generate Research Idea with Dynamic Control.

(2024).
22. T. Afonja, I. Sheth, R. Binkyte, W. Hanif, T. Ulas, M. Becker, M. Fritz, LLM4GRN: Discovering Causal Gene Regulatory Networks with LLMs -- Evaluation through Synthetic Data Generation. (2024).
23. A. Ghafarollahi, M. J. Buehler, SciAgents: Automating Scientific Discovery Through Bioinspired Multi-Agent Intelligent Graph Reasoning. *Adv. Mater.* **37**, 2413523 (2025).
24. G. Xiong, E. Xie, A. H. Shariatmadari, S. Guo, S. Bekiranov, A. Zhang, Improving Scientific Hypothesis Generation with Knowledge Grounded Large Language Models. (2024).
25. C. Si, D. Yang, T. Hashimoto, "Can LLMs Generate Novel Research Ideas? A Large-Scale Human Study with 100+ NLP Researchers" in *International Conference on Learning Representations*, Y. Yue, A. Garg, N. Peng, F. Sha, R. Yu, Eds. (2025; https://proceedings.iclr.cc/paper_files/paper/2025/file/ea94957d81b1c1caf87ef5319fa6b467-Paper-Conference.pdf)vol. 2025, pp. 94003–94092.
26. Q. Wang, D. Downey, H. Ji, T. Hope, "{S}ci{MON}: Scientific Inspiration Machines Optimized for Novelty" in *Proceedings of the 62nd Annual Meeting of the Association for Computational Linguistics (Volume 1: Long Papers)*, L.-W. Ku, A. Martins, V. Srikumar, Eds. (Association for Computational Linguistics, Bangkok, Thailand, 2024; https://aclanthology.org/2024.acl-long.18/), pp. 279–299.
27. Z. Yang, X. Du, J. Li, J. Zheng, S. Poria, E. Cambria, "Large language models for automated open-domain scientific hypotheses discovery" in *Findings of the Association for Computational Linguistics: ACL 2024* (2024), pp. 13545–13565.
28. Y. Pu, T. Lin, H. Chen, PiFlow: Principle-aware Scientific Discovery with Multi-Agent Collaboration. (2025).
29. Y. Pu, T. Lin, H. Chen, Principle-Evolvable Scientific Discovery via Uncertainty Minimization. (2026).
30. R. Vasu, C. Basu, B. Dalvi Mishra, C. Sarasua, P. Clark, A. Bernstein, "{H}yp{ER}: Literature-grounded Hypothesis Generation and Distillation with Provenance" in *Proceedings of the 2025 Conference on Empirical Methods in Natural Language Processing*, C. Christodoulopoulos, T. Chakraborty, C. Rose, V. Peng, Eds. (Association for Computational Linguistics, Suzhou, China, 2025; https://aclanthology.org/2025.emnlp-main.1292/), pp. 25413–25438.
31. Z. Yang, W. Liu, B. Gao, T. Xie, Y. Li, W. Ouyang, S. Poria, E. Cambria, D. Zhou, "{MOOSE}-Chem: Large Language Models for Rediscovering Unseen Chemistry Scientific Hypotheses" in *The Thirteenth International Conference on Learning Representations* (2025; https://openreview.net/forum?id=X9OfMNNepI).
32. Y. Liu, Z. Yang, T. Xie, J. Ni, B. Gao, Y. Li, S. Tang, W. Ouyang, E. Cambria, D. Zhou, ResearchBench: Benchmarking LLMs in Scientific Discovery via Inspiration-Based Task Decomposition. (2025).
33. Scientific Explanation. https://plato.stanford.edu/archives/win2019/entries/scientific-explanation/.
34. C. G. Hempel, P. Oppenheim, Studies in the Logic of Explanation. *Philos. Sci.* **15**, 135–175 (1948).
35. S. Yao, D. Yu, J. Zhao, I. Shafran, T. L. Griffiths, Y. Cao, K. Narasimhan, Tree of Thoughts: Deliberate Problem Solving with Large Language Models. *Adv. Neural Inf. Process. Syst.* **36**, 1–

14 (2023).
36. I. D. Cooper, How to write an original research paper (and get it published). (2015). https://doi.org/10.3163/1536-5050.103.2.001.
37. L. B. Sollaci, M. G. Pereira, The introduction, methods, results, and discussion (IMRAD) structure: a fifty-year survey. *J. Med. Libr. Assoc.* **92**, 364–367 (2004).
38. R. Arp, B. Smith, A. D. Spear, *Building Ontologies with Basic Formal Ontology* (The MIT Press, 2015; http://www.jstor.org/stable/j.ctt17kk7vw).
39. B. Smith, CLASSIFYING PROCESSES: AN ESSAY IN APPLIED ONTOLOGY. *Ratio* **25**, 463–488 (2012).
40. J. Li, H. Yu, X. Luo, Q. Liu, "{COSIGN}: Contextual Facts Guided Generation for Knowledge Graph Completion" in *Proceedings of the 2024 Conference of the North American Chapter of the Association for Computational Linguistics: Human Language Technologies (Volume 1: Long Papers)*, K. Duh, H. Gomez, S. Bethard, Eds. (Association for Computational Linguistics, Mexico City, Mexico, 2024; https://aclanthology.org/2024.naacl-long.93/), pp. 1669–1682.
41. S. Toro, A. V Anagnostopoulos, S. M. Bello, K. Blumberg, R. Cameron, L. Carmody, A. D. Diehl, D. M. Dooley, W. D. Duncan, P. Fey, P. Gaudet, N. L. Harris, M. P. Joachimiak, L. Kiani, T. Lubiana, M. C. Munoz-Torres, S. O'Neil, D. Osumi-Sutherland, A. Puig-Barbe, J. T. Reese, L. Reiser, S. M. C. Robb, T. Ruemping, J. Seager, E. Sid, R. Stefancsik, M. Weber, V. Wood, M. A. Haendel, C. J. Mungall, Dynamic Retrieval Augmented Generation of Ontologies using Artificial Intelligence (DRAGON-AI). *J. Biomed. Semantics* **15**, 19 (2024).
42. J. Gu, X. Jiang, Z. Shi, H. Tan, X. Zhai, C. Xu, W. Li, Y. Shen, S. Ma, H. Liu, S. Wang, K. Zhang, Z. Lin, B. Zhang, L. Ni, W. Gao, Y. Wang, J. Guo, A survey on LLM-as-a-judge. *Innov.* **7**, 101253 (2026).
43. D. Li, B. Jiang, L. Huang, A. Beigi, C. Zhao, Z. Tan, A. Bhattacharjee, Y. Jiang, C. Chen, T. Wu, K. Shu, L. Cheng, H. Liu, "From Generation to Judgment: Opportunities and Challenges of {LLM}-as-a-judge" in *Proceedings of the 2025 Conference on Empirical Methods in Natural Language Processing*, C. Christodoulopoulos, T. Chakraborty, C. Rose, V. Peng, Eds. (Association for Computational Linguistics, Suzhou, China, 2025; https://aclanthology.org/2025.emnlp-main.138/), pp. 2757–2791.
44. Z. Yue, H. Zeng, L. Shang, Y. Liu, Y. Zhang, D. Wang, "Retrieval Augmented Fact Verification by Synthesizing Contrastive Arguments" in *Proceedings of the 62nd Annual Meeting of the Association for Computational Linguistics (Volume 1: Long Papers)*, L.-W. Ku, A. Martins, V. Srikumar, Eds. (Association for Computational Linguistics, Bangkok, Thailand, 2024; https://aclanthology.org/2024.acl-long.556/), pp. 10331–10343.
45. A. Crisan, B. Fiore-Gartland, M. Tory, Passing the Data Baton : A Retrospective Analysis on Data Science Work and Workers. *IEEE Trans. Vis. Comput. Graph.* **27**, 1860–1870 (2021).
46. F. R. Giordano, M. D. Weir, *A First Course in Mathematical Modeling / Frank R. Giordano, Maurice D. Weir.* (Brooks/Cole Pub. Co., Monterey, CA, 1985).
47. Glenn Ledder, *Mathematics for the Life Sciences* (Springer New York, NY, 2016).
48. Law (principle). https://en.wikipedia.org/wiki/Law_(principle).
49. K. He, X. Zhang, S. Ren, J. Sun, "Deep Residual Learning for Image Recognition" in *2016 IEEE Conference on Computer Vision and Pattern Recognition (CVPR)* (2016), pp. 770–778.
50. A. Vaswani, N. Shazeer, N. Parmar, J. Uszkoreit, L. Jones, A. N. Gomez, Ł. Kaiser, I. Polosukhin, "Attention is All you Need" in *Advances in Neural Information Processing Systems*, I. Guyon, U.

Von Luxburg, S. Bengio, H. Wallach, R. Fergus, S. Vishwanathan, R. Garnett, Eds. (Curran Associates, Inc., 2017; https://proceedings.neurips.cc/paper_files/paper/2017/file/3f5ee243547dee91fbd053c1c4a845aa-Paper.pdf)vol. 30.

51. T. Mikolov, K. Chen, G. S. Corrado, J. Dean, “Efficient Estimation of Word Representations in Vector Space” in *International Conference on Learning Representations* (2013; https://api.semanticscholar.org/CorpusID:5959482).
52. arXiv.org submitters, arXiv Dataset. Kaggle (2024). https://doi.org/10.34740/KAGGLE/DSV/7548853.
53. M. Sampson, L. Zhang, A. Morrison, N. J. Barrowman, T. J. Clifford, R. W. Platt, T. P. Klassen, D. Moher, An alternative to the hand searching gold standard: validating methodological search filters using relative recall. *BMC Med. Res. Methodol.* **6**, 33 (2006).
54. Amdahl’s Law. https://en.wikipedia.org/wiki/Amdahl%27s_law.
55. Zipf’s law. https://en.wikipedia.org/wiki/Zipf%27s_law.
56. R. F. Woolson, “Wilcoxon Signed-Rank Test” in *Wiley Encyclopedia of Clinical Trials* (John Wiley & Sons, Ltd, 2008; https://onlinelibrary.wiley.com/doi/abs/10.1002/9780471462422.eoct979), pp. 1–3.
57. Salvatore S. Mangiafico, “Scheirer–Ray–Hare Test” in *Summary and Analysis of Extension Program Evaluation in R* (2016; https://rcompanion.org/handbook/F_14.html).
58. M. S. Nixon, A. S. Aguado, “12 - Distance, classification and learning” in *Feature Extraction and Image Processing for Computer Vision (Fourth Edition)* (Academic Press, Fourth Edi., 2020; https://www.sciencedirect.com/science/article/pii/B9780128149768000129), pp. 571–604.
59. Y.-H. H. Tsai, S. Bai, M. Yamada, L.-P. Morency, R. Salakhutdinov, “Transformer Dissection: An Unified Understanding for Transformer{’}s Attention via the Lens of Kernel” in *Proceedings of the 2019 Conference on Empirical Methods in Natural Language Processing and the 9th International Joint Conference on Natural Language Processing (EMNLP-IJCNLP)*, K. Inui, J. Jiang, V. Ng, X. Wan, Eds. (Association for Computational Linguistics, Hong Kong, China, 2019; https://aclanthology.org/D19-1443/), pp. 4344–4353.
60. P. N. Swarztrauber, On Computing the Points and Weights for Gauss--Legendre Quadrature. *SIAM J. Sci. Comput.* **24**, 945–954 (2003).
61. P. S. Heckbert, “Fourier Transforms and the Fast Fourier Transform ( FFT ) Algorithm” (1998; https://api.semanticscholar.org/CorpusID:6022157).
62. A. Katharopoulos, A. Vyas, N. Pappas, F. Fleuret, “Transformers are RNNs: fast autoregressive transformers with linear attention” in *Proceedings of the 37th International Conference on Machine Learning* (JMLR.org, 2020)*ICML’20*.
63. Y. Chen, K. Ren, Y. Wang, Y. Fang, W. Sun, D. Li, “ContiFormer: continuous-time transformer for irregular time series modeling” in *Proceedings of the 37th International Conference on Neural Information Processing Systems* (Curran Associates Inc., Red Hook, NY, USA, 2023)*NIPS ’23*.
64. A. Baevski, M. Auli, “Adaptive Input Representations for Neural Language Modeling” in *International Conference on Learning Representations* (2019; https://openreview.net/forum?id=ByxZX20qFQ).
65. T. Mikolov, I. Sutskever, K. Chen, G. Corrado, J. Dean, “Distributed representations of words and phrases and their compositionality” in *Proceedings of the 27th International Conference* on Neural Information Processing Systems - Volume 2 (Curran Associates Inc., Red Hook, NY, USA, 2013)NIPS’13, pp. 3111–3119.
66. Y. Pinter, R. Guthrie, J. Eisenstein, “Mimicking Word Embeddings using Subword {RNN}s” in *Proceedings of the 2017 Conference on Empirical Methods in Natural Language Processing*, M. Palmer, R. Hwa, S. Riedel, Eds. (Association for Computational Linguistics, Copenhagen, Denmark, 2017; https://aclanthology.org/D17-1010/), pp. 102–112.
67. R. Shu, H. Nakayama, “Compressing Word Embeddings via Deep Compositional Code Learning” in *International Conference on Learning Representations* (2018; https://openreview.net/forum?id=BJRZzFlRb).
68. P. Bojanowski, E. Grave, A. Joulin, T. Mikolov, Enriching Word Vectors with Subword Information. *TACL* **5**, 135–146 (2017).
69. D. Svenstrup, J. M. Hansen, O. Winther, “Hash embeddings for efficient word representations” in *Proceedings of the 31st International Conference on Neural Information Processing Systems* (Curran Associates Inc., Red Hook, NY, USA, 2017)*NIPS’17*, pp. 4935–4943.
70. G. E. Karniadakis, I. G. Kevrekidis, L. Lu, P. Perdikaris, S. Wang, L. Yang, Physics-informed machine learning. *Nat. Rev. Phys.* **3**, 422–440 (2021).

# Appendix A. Continuous-Time Attention Transformer

## A.1 Overall Architecture

CTAT (Continuous-Time Attention Transformer) Models: Instead of using positional encoding and content-independent dot-product bias for modeling relative positional distance, we used a Gaussian distance kernel. Further, we tested four different computation methods (where $L$ denotes the sequence length):

1) Dense Softmax, which explicitly computes $O(L^2)$ attention scores and adds a Gaussian distance kernel bias.
2) Dense Linear (ELU+1), which is a linear form of attention that explicitly computes the kernel-weighted sum in $O(L^2)$.
3) FFT Linear, which uses the convolution theorem and a Fast Fourier Transform to reduce the complexity of the linear attention to $O\big(L\,log(L)\big)$.
4) Gauss-Legendre finite-interval approximation, which approximates the continuous-time integral using a fixed number of quadrature nodes, but is restricted to a learnable causal window, where $O(LM), M$ is the number of interpolation nodes.

## A.2 Key Mathematical Definitions

### A.2.1 Gaussian Distance Kernel

For positions $i$ and $j (j \le i)$, the time distance is $r = i - j \ge 0$. Each attention head $h$ has an inverse bandwidth parameter $m_h > 0$. The Gaussian distance kernel is defined as:

$$K_h(r) = exp(-\frac{1}{2}(m_h r)^2) \tag{7}$$

Here,

- a large $m_h$ leads to a shorter memory (i.e., the kernel decays faster), and vice versa.
- $m_h$ is a learnable parameter.

### A.2.2 Standard Softmax with Gaussian Distance Bias (Gaussian Relative Position Encoding)

In standard scaled dot-product attention, the Gaussian kernel function can be merged with the exponent. Thus, the logits are expressed as:

$$logits_{i,j,h} = \frac{\boldsymbol{q}_{i,h}^T \boldsymbol{k}_{j,h}}{\sqrt{d_k}} - \frac{1}{2}\left(m_h(i-j)\right)^2 \tag{8}$$

where $\boldsymbol{q}_{i,h}^T, \boldsymbol{k}_{j,h}$ are in bold to denote vectors or matrices. Then causal masking ($j \le i$) and softmax are applied, and the result is multiplied by $\boldsymbol{v}$. The complexity is $O(L^2)$.

### A.2.3 Linear Attention with Gaussian Distance Bias

The core idea of linear attention is to transform $\boldsymbol{q}, \boldsymbol{k}$ into non-negative features using a non-linear map $\phi(x) = \text{ELU(x)} + 1$, and then approximate a softmax function via a kernel decomposition of softmax such as $\phi(\boldsymbol{q})^T \phi(\boldsymbol{k})$.

$$\phi(x) = \text{ELU(x)} + 1 = \begin{cases} \text{x} + 1 & \text{if x} > 0 \\ \text{exp(x)} & \text{if x} \le 0 \end{cases} \tag{9}$$

The output of causal linear attention would then be:

$$y_i = \frac{\sum_{j \le i}\left(K(i-j)\phi(\boldsymbol{q})_i^T \phi(\boldsymbol{k})_j\right)\boldsymbol{v}_j}{\sum_{j \le i}\left(K(i-j)\phi(\boldsymbol{q})_i^T \phi(\boldsymbol{k})_j\right)} \tag{10}$$

$$y_i = \frac{\left(\phi(\boldsymbol{q})_i^T \sum_{j \le i} K(i-j)\left(\phi(\boldsymbol{k})_j v_{j\ j}^T\right)\right)^T}{\phi(\boldsymbol{q})_i^T \sum_{j \le i}\left(K(i-j)\phi(\boldsymbol{k})_j\right)} \tag{11}$$

$$\mathrm{A_i} = \sum_{j \le i} K(i-j)\left(\phi(\boldsymbol{k})_j v_{j\ j}^T\right) \tag{12}$$

$$\mathrm{z_i} = \sum_{j \le i}\left(K(i-j)\phi(\boldsymbol{k})_j\right) \tag{13}$$

Dense Linear: builds the $L \times L$ weight matrix explicitly.
FFT computation: Eq. (10) can be transformed into Eq. (11) because the numerator in Eq. (10) has causal convolution. $\mathrm{A_i}$ in Eq.(12) and the denominator in Eq.(11) has causal convolution. $\mathrm{z_i}$ in Eq. (13) can therefore be computed by the convolution property. Because the kernel only depends on the time difference between $(i-j)$, these sums are causal convolutions, that can be implemented in $O\big(L\,log(L)\big)$ using a Fast Fourier Transform.

## A.2.3 Continuous-Time Interpretation and Numerical Integration Approximation (CTAT)

Think of discrete token indices as equally spaced points on a continuous time axis $t_i = \mathrm{i}$. Then write the causal attention as a continuous integral:

$$y(t_i) = \frac{\int_0^{t_i} K(t_i-\tau) exp(\frac{\boldsymbol{q}(t_i)^{\boldsymbol{T}}\boldsymbol{k}(\tau)}{\sqrt{d_k}})\boldsymbol{v}(\tau)\, d\tau}{\int_0^{t_i} K(t_i-\tau) exp(\frac{\boldsymbol{q}(t_i)^{\boldsymbol{T}}\boldsymbol{k}(\tau)}{\sqrt{d_k}})\, d\tau} \tag{14}$$

A numerical integration with $\boldsymbol{M}$ nodes approximates the integral:

$$\int_0^{t_i} f(\tau)\, d\tau \approx \sum_{i=1}^{M} w_{i,m} f(\tau_{i,m}) \tag{15}$$

The sample points $\tau_{i,m}$ and $w_{i,m}$ are determined by orthogonal polinomials. $\boldsymbol{k}\big(\tau_{i,m}\big)$ **and** $\boldsymbol{v}(\tau_{i,m})$ are obtained by linear interpolation between the two nearest integer indices. This approach reduces the complexity to $O(LM)$ because each postion $i$ only needs to evaluate $M$ sample points.

- **Gauss-Legendre Finite-interval Approximation (GLD):**

Instead of assuming $t_i$ is infinite, the integral is restricted to a learnable causal window $W_h$ (or the full $[0, t_i]$):

$$\int_0^{min(t_i,W_h)} K(m_h\tau) f(\tau)\, d\tau \tag{16}$$

Linearly map the interval to $[-1,1]$, $l_i = min(t_i, W_h)$. Define $u = \frac{2\tau-(0+l_i)}{l_i-0}$ to map to$[-1,1]$, $\tau = \frac{ul_i+l_i}{2}$,

$$\int_0^{min(t_i,W_h)} K(m_h\tau) f(\tau)\, d\tau \tag{17}$$

$$=\frac{l_i}{2}\int_{-1}^{1} K\left(m_h\left(\frac{ul_i+l_i}{2}\right)\right) f(\frac{ul_i+l_i}{2})\, du$$

$$\approx \frac{l_i}{2}\sum_{n=1}^{n=M} w_n K\left(m_h\left(\frac{l_i x_n + l_i}{2}\right)\right) f(\frac{l_i x_n + l_i}{2})$$

where $x_n$ are the Legendre interpolation nodes. Notably, GLD has a self-anchor, the current moment $t_i$ itself is also added as a sample point with a fixed weight (=1.0) to ensure the model can always attend to the present.

On the linear interpolation for the fractional position, the sampling points $\tau_{i,m}$ generated by the Gauss-Legendre quadrature are general fractional time indices. Our available points are all integer time indices for Transformer. Hence, we have used linear interpolation between the two nearst integer indices to approximate the fractional position. For each element $p$, the two neighbouring integer indices are

$$\ell = \lfloor p \rfloor, r = \lceil p \rceil \tag{18}$$

The fractional part is

$$\alpha = p - \ell \in [0.1) \tag{19}$$

The linearly interpolated vector is

$$\hat{z} = (1-\alpha) z[\ell] + \alpha z[r] \tag{20}$$

where $z[i]$ is the sampled values retured by the integral function.

# Appendix B. Heaps-Adaptive Lexico Manifold Embedding (HALO)

## B.1 Overall Architecture

HALO is a word embedding model that decomposes each word vector into two complementary parts:

- An explicit core embedding for high-frequency words (stored as a classic lookup table).
- An implicit subword/manifold embedding generated from character n-grams, shared across all words.

A soft, frequency-aware gate then blends these two components per word. The gate is guided by a prior derived from word frequency that is partially learnable. This design aims to retain the precision of dedicated embeddings for frequent words while enabling robust generalization for rare and unseen words via subword information – all with a parameter count that is largely decoupled from vocabulary size.

The training objectives built on top of this embedding are:

- HALO-SGNS: Skip-gram with negative sampling using HALO embeddings.
- HALO-CBOW: CBOW with negative sampling using HALO embeddings.

## B.1 Explicit Core Embeddings

We split the vocabulary $V$ (size $|V|$) into a core set of the $C$ most frequent words and a remaining set containing all other words, where $C \ll |V|$.
We then defined a dedicated embedding matrix as follows:

$$E^{exp} \in R^{(C+1)d}$$

- Row 0 is permanently fixed to 0 and serves as the representation for non-core words.
- Row $k\ (1 \leq k \leq C)$ stores the explicit $d$-dimensional vector for the $k$-th core word.

The core index $I(j)$ of a word with a global vocabulary index $j$, is defined as:

$$I(j) = \begin{cases} 0, & if\ word\ j\ is\ not\ a\ core\ word \\ k, & if\ the\ word\ j\ is\ the\ k-th\ core\ word \end{cases}$$

The explicit embedding of word $j$ is then simply a lookup:

$$e_j^{exp} = E_{I(j)}^{exp} \in R^d$$

Non-core words receive the zero vector 0. The number of parameters in this component is $(C+1)d$, which only grows with the chosen core size, not with the total vocabulary $|V|$. This significantly reduces the vocabulary parameters.

## B.2 Implicit Subword/Manifold Path

The implicit path produces a continuous vector for every word based solely on character-level features and a compact set of shared parameters. It does not explicitly store any per-word vectors.

### B.2.1 Character n-gram hashing

Given a word $w$, we first add boundary markers ('<' and '>') which denote the start and end points for extracting n-grams of a length $n = n_{min}, n_{min}+1, n_{min}+2, \ldots n_{max}$. Each distinct n-gram is mapped to a bucket ID using a deterministic hash function $H$:

$$b = H(n_gram) \in \{1,2,\ldots,B\}$$

where $B$ is the total number of subword buckets.
For each word, the first $M$ bucket IDs are kept and padded with zeros if the word has fewer than $M$ n-grams (where $M$ is the max number of features). This yields a fixed-length feature vector for word $j$ of：

$$f_j = (f_{j,1}, f_{j,2}, \ldots, f_{j,M}), f_{jt} \in \{0,1,\ldots,B\}$$

### B.2.2 Subword embedding aggregation

A shared subword embedding table is defined:

$$S \in R^{(B+1)D_s}$$

where row 0 is fixed to zero (padding), and $D_s$ is the subword dimension.

For word j, the embeddings of its n-gram are $\{s_{f_{j,t}}\}_{t=1}^{M}$ with $s_0 = 0$.

These are averaged over the valid n-grams to obtain a fixed-size subword summary vector:

$$h_j = \frac{\sum_{t=1}^{M} s_{f_{j,t}} \cdot 1[f_{j,t} \neq 0]}{max(1, \sum_{t=1}^{M} 1[f_{j,t} \neq 0])} \in R^{D_s}$$

Since the feature extraction and averaging are deterministic, $h_j$ can be computed for any word, including out of vocabulary words, as long as its n-grams can be hashed into the existing buckets.

### B.2.3 Manifold projection and residual

The vector $h_j$ is transformed into a full $d$-dimensional embedding using two complementary pathways.

(a) Low-rank manifold combination:

A learnable basis matrix is stored:

$$B \in R^{K \times d}$$

where $K$ is the number of bases. The rows of $B$ span a low-dimensional “word manifold” in the embedding space.

From $h_j$, we compute the coordinates on this manifold through a linear layer followed by tanh:

$$a_j = tanh(W_{coeff} h_j + b_{coeff}) \in R^K$$

where $W_{coeff} \in R^{K \times D_s}$ and $b_{coeff} \in R^K$ are trainable parameters. The manifold component is then:

$$m_j = a_j B = \sum_{k=1}^{K} a_{j,k} B_{k,:} \in R^d$$

(b) Residual path

A direct linear projection provides a full-dimensional residual vector:

$$r_j = \rho \cdot (W_{res} h_j + b_{res}) \in R^d$$

with $W_{res} \in R^{d \times D_s}$, $b_{res} \in R^d$, and a scalar $\rho$ that controls the contribution

(c) Final implicit embedding

The implicit representation of word $j$ is the sum of the two paths:

$$z_j = m_j + r_j$$

The total number of parameters in this entire implicit path is bounded by the dimensions of the subword table, the linear layers, and the basis matrix. This makes the model parameter-efficient and allows it to generalize to unseen words.

## B.3 Frequency-Aware Soft Gating and Auxiliary Losses to Prevent Gate Collapse

The HALO embedding blends the explicit core vector and the implicit subword vector through a soft, frequency-guided gate. The gate is equipped with auxiliary losses that prevent it from collapsing to all-implicit behavior.

### B.3.1 Soft gating function

For each word $j$, let $f_j$ be its corpus frequency (or its unigram count). Then, we define the smoothed log-frequency:

$$l_j = log(f_j + 1)$$

and introduce a learnable scalar threshold $\gamma$ at a temperature of $\tau > 0$.
For a core word, the gate value is:

$$g_j = \sigma(\frac{l_j - \gamma}{\tau})$$

where $\sigma(x) = {}^{1}/_{(1+e^{-x})}$ is the sigmoid function.

For a non-core word, we simply set $g_j = 0$ (masking it with a binary indicator $\delta_j^{core}$ if the word $j$ is a non-core word). The initial value of $\gamma$ is set to $\gamma_0 = l_{core_min}$, which is the log-frequency of the least frequent word among the $C$ core words. This places the transition region right at the frequency cutoff that separates core words from non-core words.
The final embedding of word $j$ is :

$$v_j = g_j \cdot e_j^{exp} + (1 - g_j) \cdot z_j$$

Thus, very frequent core words ($l_j - \gamma \gg 0$) have $g_j \to 1$ and are almost entirely explicit vectors. Core words below the threshold (including non-core words) only use the implicit path. $\gamma$ is a learnable parameter that allows the model shift the transition point based on training evidence, while $\tau$ controls the smoothness of the transition.

### B.3.2 Auxiliary Losses to Prevent Gate Collapse

Relying only on the main training objective (Skip-gram or CBOW negative sampling) can cause the gate to degenerate in that $\gamma$ might drift to extreme values, making all core gates nearly 0 or nearly 1. This would destroy the intended frequency-adaptive blend. To counteract this, we added two auxiliary loss terms, with the weights $\lambda_{exp}$, $\lambda_{prior}$.
(a) Explicit core loss $L_{exp}$(weight $\lambda_{exp}$)
This loss is computed exclusively through the explicit embedding pathways (using $e^{exp}$ on both the input and output sides) and is applied only to examples where all words involved (center, positive context/target, and negative samples) are core words.
For the CBOW case, this is expressed as:

$$L_{exp} = -\log\sigma(\bar{v}^{exp} \cdot u_t^{exp}) - \sum_{n\in N} \log\sigma(\bar{v}^{exp} \cdot u_n^{exp})$$

where

$$\bar{v}^{exp} = \frac{\sum_{i=1}^{k} m_i \cdot e_{c_i}^{exp}}{\max(\sum_{i=1}^{k} m_i, 1)}$$

$m_i = 1$ when the $i$th context word is an explicit word; otherwise, $m_i = 0$. $u_t^{exp}$ is the positive context word, while $u_n^{exp}$ is negative context word.

(b) Gate prior loss $L_{prior}$(weight $\lambda_{prior}$)
We define a fixed prior gate using the initial threshold $\gamma_0$ and a separate temperature $\tau_{prior}$:

$$g_j^{prior} = \sigma\left(\frac{l_j - \gamma_0}{\tau_{prior}}\right) \cdot \delta_j^{core}$$

$\delta_j^{core}$ sets the implicit words to 0.
The current gate $g_j$ (from the main forward pass) is penalized if it deviates from this prior via the mean squared error:

$$L_{prior} = \frac{1}{|S|}\sum_{j \in S}\left(g_j - g_j^{prior}\right)^2$$

where $S$ is the set of words seen in the current batch. This term acts as soft anchor, pulling the learnable $\gamma$ back to wards its initial value and keeping the gate consistent with the raw frequency ordering.

(c) The full training objective
Let $L_{mix}$ be the standard negative-sampling loss applied to the blended HALO embeddings (the main objective). The total loss for HALO-SGNS or HALO-CBOW is:

$$L = L_{mix} + \lambda_{exp} L_{exp} + \lambda_{prior} L_{prior}$$

This formulation allows HALO to a learn a frequency-adaptive decomposition where high-frequency words retain a dedicated vector (explicit core), mid-range frequency words smoothly transition to using the manifold information, and rare and unseen words rely entirely on the subword-driven implicit path.

# Appendix C. Prompt Template of DN Decomposition for data science modelling

| Prompt Template of DN Decomposition for data science modelling |
|---|
| **Answer Template**:<br>###DN_START###<br><br>## [EXPLANANDUM_START]<br>>> [Explanandum:... ]<br>## [EXPLANANDUM_END]<br><br>## [GENERAL_LAWS_START]<br>* Law 1 ([Law Name]): [Description of the first general scientific law or principle relevant to the explanation.] |

* Law 2 ([Law Name]): [Description of the second general law.]
* Law 3 ([Law Name]): [Description of the third law.]
* ... [Add or remove as needed]
## [GENERAL_LAWS_END]

## [INITIAL_CONDITIONS_START]
* Condition 1: [Statement of the first specific initial condition, fact, or setting in this particular research.]
* Condition 2: [Statement of the second condition.]
* Condition 3: [Etc.]
## [INITIAL_CONDITIONS_END]

## [DEDUCTIVE_REASONING_START]
1. [First step in the logical deduction connecting laws and conditions to the result.]
2. [Second logical step.]
3. [...]

[Final integrative or summary reasoning, if any.]
## [DEDUCTIVE_REASONING_END]

## [TABLE_START]
| Law 1: [Short Law 1 summary]; | | |
| Law 2: [Short Law 2 summary]; | | |
| Condition 1: [Short Condition 1 summary]; | | |
| Condition 2: [Short Condition 2 summary]; | | |
| ... | | |
| [Optionally, summarize main conclusion here] | | |
## [TABLE_END]

## [CONCLUSION_START]
[Concluding sentence or two summarizing how the laws and conditions explain the phenomenon.]
## [CONCLUSION_END]

###DN_END###
**Template end**

# Appendix D. Prompt for Law Evaluation in Table 1

Prompt for Law Evaluation in Table 1

You are a top-tier strategic evaluation expert in AI research. The user now has a clear core phenomenon or requirement: {core_demand} (e.g., improving the reasoning performance of large language models, enhancing the accuracy of vision models, overcoming bottlenecks in long-context modeling for text classification, etc.).

You will use the "Four-Dimensional Potential Assessment Framework" to systematically score each law/theory/principle provided by the user.
The assessment framework strictly uses the following 4 dimensions, each scored from 0 to 10, with a maximum total score of 40:

1. Relevance
10 = The law is highly and directly relevant to the target phenomenon; its core mechanism naturally aligns with the key driving factors of the phenomenon;
0 = Almost irrelevant or merely a superficial analogy; unable to establish a meaningful causal connection.

2. Gap
10 = The law has not yet been explicitly proposed or systematically applied to explain this phenomenon, indicating a significant knowledge/methodological gap;
0 = The law has been extensively used to explain this phenomenon; the theory is saturated with no room for new insights.

3. Logic
10 = When applying the law to the phenomenon, the reasoning chain is clear and free of internal contradictions, and it is compatible with existing mainstream frameworks or provides a superior explanation;
0 = Logical breaks, circular reasoning, or conflicts with established facts; unable to form a self-consistent explanatory system.

4. Feasibility
10 = Under existing or near-future experimental/model settings, clear quantitative or qualitative verification schemes can be designed (e.g., comparative experiments, ablation studies, observable metrics);
0 = Verification requires unattainable technological breakthroughs, infinite computational resources, or unobservable assumptions; practically infeasible.

Output Format (strictly follow):

| Name of Law/Theory/Principle | Relevance | Gap | Logic | Feasibility | Total Score | Potential Ranking for the Phenomenon (2025 – 2028) | One-Sentence Matching Summary |

|---------------------------|-----------|-----|-------|-------------|------|------------------------------------|-----------------|
| ... | ... | ... | ... | ... | ... | ... | ... |

After the table, write a summary of about 200 words in Chinese, clearly identifying the top 1–3 laws most worth going all-in on right now. For each, provide a specific, high-impact innovation roadmap (50–100 words each), focusing on how to leverage its theoretical gap and logical advantages to design verifiable experiments or model architectures.

After the summary, leave two blank lines, and finally output the total scores sorted from highest to lowest in the following fixed format (must be exactly the same, no more, no less):
[Total Score Ranking]
Law Name: XXX1 | Total Score: yy
Law Name: XXX2 | Total Score: yy
Law Name: XXX3 | Total Score: yy
(Only list laws with scores, in descending order of total score)

Now, begin evaluating all the laws in the following JSON list:

# Appendix E. Prompt for Hypothesis Evaluation

Prompt for Hypothesis Evaluation in Table 3

You are a harsh reviewer in Computer Science. Evaluate the research hypothesis from four aspects on a 5-point scale.

You must output a valid JSON object with the following keys: "Validness", "Novelty", "Significance", "Potential".
For each key, the value is an object containing:
"score": an integer from 1 to 5,
"reason": a string explaining the score.
Do not include any additional text, explanation, or formatting outside the JSON object.

```
{
  "Validness": {
      "score": null,
      "reason": ""
    },
    "Novelty": {
      "score": null,
      "reason": ""
    },
```

```
    "Significance": {
      "score": null,
      "reason": ""
    },
    "Potential": {
      "score": null,
      "reason": ""
    }
}
```

Scoring criteria:
Aspect 1: Validness
5: Strongly grounded in theory, prior work, or preliminary evidence; implementation and evaluation are clearly feasible with available methods.
4: Logically sound and well supported by existing knowledge; evaluation path is clear and practical.
3: Plausible but uncertain; relies on assumptions that are reasonable but not yet well validated.
2: Weakly supported or difficult to evaluate; key assumptions, mechanisms, or feasibility are unclear.
1: Contradicts established principles, lacks a coherent rationale, or cannot be realistically implemented or evaluated.

Aspect 2: Novelty
5: Introduces a fundamentally new problem, method, abstraction, or perspective with little direct precedent.
4: Substantially departs from existing work by changing key assumptions, formulations, or techniques.
3: Provides a clear new idea within an established research framework.
2: Mostly recombines or slightly modifies known ideas, with limited conceptual originality.
1: Trivial extension, direct application, or near-duplicate of prior work.

Aspect 3: Significance
5: If successful, could substantially change a major area of computer science or enable capabilities not currently possible.
4: Could resolve an important limitation, challenge prevailing assumptions, or produce broad advances in a field.
3: Could deliver meaningful improvement for a specific research area, task, or class of systems.
2: Provides limited or incremental benefit in a narrow setting.
1: Has little practical, theoretical, or scientific importance beyond a very specific case.

Aspect 4: Potential
5: Opens a broad, self-contained research agenda with many natural follow-up questions, benchmarks, methods, or applications.
4: Creates a promising research direction that could support multiple follow-up studies by different groups.
3: Suggests some follow-up work, but the resulting research space is moderate in scope.
2: Follow-up work is possible but likely limited to incremental refinements or extensions.
1: Unlikely to generate further research; the idea is too narrow, closed-ended, or exhausted once tested.

The hypothesis is:
{hypothesis}

core need is:
{core_demand}

# Appendix F. Prompt for Process modeling by BFO

You are an ontology engineering assistant following Basic Formal Ontology (BFO) and an ontological realism perspective.
Given a description of a scientific phenomenon and its formation process, your task is to:
1. Perform a BFO-style process ontology modeling of that phenomenon.
2. Produce a BFO-style list of types/universals involved in that process.

Follow these rules carefully:

---

1. Ontological stance
- Adopt an ontological realist view.
- Use "type/universal" to refer to repeatable entities: natural kinds, process types, attribute types, role types, etc., that can be instantiated multiple times in reality.
- Use "individual/instance" to refer to particular, single, concrete entities or events in space and time.
- Do NOT output individuals. Only output types/universals.

---

2. Process modeling tasks
For the input process description, do the following (and show the results explicitly):
1. Top-level process universal
   o Identify the phenomenon as a process universal.
   o Model it as a subclass of bfo:process (or equivalent).
   o Provide a concise Aristotelian-style definition:
     genus: process
     differentia: what makes this process type distinctive (participants, mechanism, outcome, conditions).
2. Subprocess universals

o Decompose the top-level process into key subprocess types (stages, phases, mechanisms).
o Each subprocess is also a subclass of process.
o Indicate their ordering or structural relations using has_part, has_participant, precedes, or similar (natural language is fine).
3. Participant universals (continuants)
o List all relevant continuant types involved:
e.g. material entities, organisms, organs, cells, molecules, instruments, celestial bodies, etc.
o Only include repeatable kinds, not specific samples or named individuals.
4. Role universals
o Identify roles realized in the process:
e.g. reactant role, product role, catalyst role, signal role, carrier role, source role, sink role.
o Model them as role universals (realizable entities).
o For each role, specify:
its typical bearer type (which entity type can bear this role),
in which process type it is realized.
5. Quality universals
o Identify relevant qualities:
e.g. mass, charge, temperature, pressure, pH, density, velocity.
o For each, indicate:
which entity types they inhere in,
how they are relevant to the process.
6. Disposition universals
o Identify relevant dispositions / propensities:
e.g. radioactivity, flammability, solubility, catalytic activity, excitability, binding affinity.
o For each, indicate:
its bearer type,
the process type in which it is realized.
7. Input / output universals
o Identify input types:
substances, entities, energy forms, signals consumed or required by the process.
o Identify output types:
products, transformed entities, emitted radiation, byproducts, structural changes.
o Model them as material/entity types or appropriate universals.
8. Sites / spatial and structural context universals
o Where relevant, specify site / spatial region / structural context types:
e.g. intracellular space, membrane region, fault zone, reaction vessel interior, boundary layer, atmosphere layer.
o These should also be universals.
9. Relational constraints (at type level)
o Use type-level relations (in natural language) to capture constraints such as:
“Every instance of [subprocess] has_participant some [entity type].”
“Every instance of [disposition] is realized_in some [process type].”
“Every instance of [role] is borne_by some [entity type].”
o Keep them clearly at the universal/type level.

———————————————————————————

3. Selection criteria for types/universals

When constructing the list:

- Include only: entities that can have multiple instances in principle (repeatable, general types).
- Exclude:
  - One-off historical events.
  - Named individual objects (e.g. “this specific volcano”, “Experiment #12”).
  - Purely ad hoc groupings defined only by arbitrary thresholds or local conventions (unless they correspond to robust scientific kinds or mechanisms).
- If a candidate is a complex logical combination (e.g. “X with property ⩾ threshold”), treat it as:
  - a defined class built from more basic universals,
  - not necessarily a fundamental universal, unless justified by stable theory.

For each listed universal, briefly note:

- its BFO upper-level category:
  - process, material entity, object, fiat object part, site,
  - quality, disposition, role, etc.
- its function or relevance within the phenomenon’s formation process.

———————————————————————————

4. Output format

Structure your answer as follows (adapt to the specific case):

1. Top-level Process Universal
   - Name:
   - BFO category: process
   - Definition:
2. Subprocess Universals
   - Bullet list. For each:
     - Name
     - BFO category: process
     - Informal description and relation to top-level process.
3. Participant Universals (Continuants)
   - Bullet list. For each:
     - Name
     - BFO category (e.g. material entity, object)
     - Role in the process (informally).
4. Role Universals
   - Bullet list. For each:
     - [Role name] — bearer: [entity type] — realized_in: [process type].
5. Quality Universals
   - Bullet list:
     - [Quality] — inheres_in: [entity type] — relevance: [short note].
6. Disposition Universals
   - Bullet list:

[Disposition] — bearer: [entity type] — realized_in: [process type].

7. Input / Output Universals
   - o Input types:
   - o Output types:
8. Sites / Spatial & Structural Context Universals
   - o Bullet list of relevant site/region/structure types.

______________________________________________

5. In case of missing details

- If the input description is incomplete, use standard scientific knowledge to propose a plausible, explicit BFO-style model.
- Clearly indicate when you are making scientifically reasonable modeling assumptions.
- Do not respond with “cannot model”; always provide the best-effort, consistent BFO-style ontology for the process.

Using the above instructions, model the formation process of [NAME OF SCIENTIFIC PHENOMENON]
and produce the corresponding BFO-style list of types/universals.

Here is the description of the phenomenon:

# Appendix G. Prompt for making assumption and modelling

prompt_assumption_modelling_input = """
You are a senior AI modeling expert. Your task is: based on the given modeling theory, directly design the most promising, novel, and implementable conceptual modeling scheme to better satisfy the core modeling requirements.

All responses must be in Chinese, and strictly follow the specified JSON format. Do not output any additional explanations or Markdown markers.

【Input Information】
Theory/Law Details:
{law}
Core Modeling Requirements:
{core_demand}
"""

prompt_assumption_modelling = """

【Core Task】

Your task is to design exactly one most promising novel conceptual modeling scheme. The scheme must:

Be grounded in the theory/law details above, and propose modeling assumptions.
Based on the assumptions, propose a fundamentally new method with no direct precedent, whose theoretical and implementation pathways are fully clear and feasible. If successful, it would be disruptive enough to revolutionize a major field of computer science, and would naturally give rise to a broad and self-consistent set of follow-up research directions.
Precisely target the core requirements stated above.
【Output Requirements】
Output exactly one valid JSON object. Do not output any additional text, comments, or code fences (such as ```json). All strings must be in Chinese and must form valid JSON strings (newlines escaped as \n).

For mathematical formula expressions:

Output pure LaTeX math source code only; do not output any explanatory text.
Formulas must not contain Chinese, Chinese punctuation, or Unicode math symbols.
Must use standard LaTeX commands, e.g., \alpha, \sum, \prod, \in, \mathbb{R}^d.
Pseudo-notation is prohibited, e.g., Σ_{i=1..m}, p_hat(w), R^d, \inD, \inNeg, \inpath(w).
Formulas must be compilable by pdfLaTeX + amsmath + amssymb + amsfonts.
Self-check and correct before output.
JSON structure:
{
"new_conceptual_modeling_scheme": {
"scheme_id": "string (e.g., S1)",
"scheme_name": "string (Chinese name of the model)",
"model_assumption": "string (modeling assumptions derived from the theory)",
"model_family": "string",
"core_conceptual_idea": "string (2-4 sentences describing the core innovative idea and intuition of the model, in Chinese)",
"modeling_translation": {
"core_structure": "string (describe only the most critical structure supporting the core concept)",
"representation_schema": "string (describe the representation form of inputs/outputs/intermediates states)",
"key_hyperparameters": ["string"],
"inference_and_objective": {
"prediction_or_scoring": "string",
"objective_function": {
"primary_criterion": "string",
"regularization_terms": ["string"]
},
"optimization_algorithm": "string"
},

"mathematical_relations": [
{
"name": "string",
"formula": "string",
"notes": "string"
}
]
},
"expected_advantages_vs_existing": {
"rationale_and_reasoning": {
"starting_assumption": "string",
"limitation_of_existing": "string",
"proposed_mechanism": "string",
"causal_chain_to_improvement": ["string"],
"summary_claim": "string"
},
"likely_improvements": ["string"],
"likely_weaknesses": ["string"]
}
}
}

For the formula field in mathematical_relations:
Output pure LaTeX math source code only
Do not output any Chinese explanations, natural language descriptions, or comments
Do not wrap with
...
,
...
,
...
, or Markdown code blocks

Do not use Unicode math symbols, e.g., α, β, Σ, ∏, ∈, ⩽, ≈, •
Must use standard LaTeX notation, e.g., \alpha, \beta, \sum, \prod, \in, \mathbb{R}^d
Pseudo-LaTeX is prohibited, e.g., Σ_{i=1..m}, p_hat(w), R^d, \inD, \inNeg, \inpath(w)
For the notes field:
Chinese explanations are allowed
Do not write explanations inside formula; put them only in notes
For narrative fields such as core_structure, representation_schema, prediction_or_scoring:
Output natural language descriptions
Inline formulas must be written as
...
Display formulas must be written as

```
...
Formulas in the narrative must also use standard LaTeX; do not use Unicode math symbols or pseudo-LaTeX
Before generating, self-check:
Does formula contain any Chinese?
Does it contain Unicode math symbols?
Does it contain pseudo-LaTeX?
Are there any unclosed \begin...\end...?
Are there any empty \text{}?
Are there any cases where \in is not followed by a set object?
If a formula lacks necessary mathematical objects and cannot be inferred:

Set formula to an empty string
Explain in notes which information is missing
Please follow the above rules strictly.

"""
```

## Appendix H. Prompt for hypothesis direct generation

```
prompt_research_idea=f"""
**Prompt for Generating Highly Innovative Research Ideas**

**System Message**:
You are an advanced AI research assistant designed to identify and generate highly innovative research ideas. Your task is to analyze the provided scientific article content and core research demand, then propose a unique, feasible, and impactful research idea.

**Input Variables**:
- **extracted_text**: {extracted_text}
- **core_demand**: {core_demand}

**Task Instructions**:
1. **Understand the Context**: Thoroughly read and understand the provided `extracted_text` to grasp the article's main contributions, methods, and findings.
2. **Identify the Gap**: Based on the `core_demand`, identify gaps, challenges, or unexplored areas within the article's scope or related fields.
3. **Generate an Innovative Idea**: Propose a research idea that is not only novel but also has the potential to significantly advance the field. The idea should directly address the identified gap and align with the `core_demand`.
4. **Provide Rationale**: Briefly explain why your proposed research idea is innovative, feasible, and important. Highlight its potential scientific, technological, or practical impact.
```

**Example Format** (for reference, not to be copied directly):

**extracted_text**:
"This study explores the application of deep learning in medical diagnosis, demonstrating high accuracy in classifying various diseases from medical images. However, challenges remain in handling rare diseases due to limited data availability."

**core_demand**:
"Improve the accuracy and efficiency of diagnosing rare diseases using advanced computational methods."

**Generated Research Idea**:
1. **Idea**: Develop a transfer learning framework that leverages large datasets of common diseases to improve the diagnosis of rare diseases by fine-tuning models on small, specialized rare disease datasets.
2. **Rationale**: This approach addresses the data scarcity issue for rare diseases by utilizing knowledge transferred from common diseases, potentially increasing diagnostic accuracy and efficiency for rare conditions, thus meeting the core demand.

**Your Turn**:
Now, please fill in the `extracted_text` and `core_demand` with the actual content from the article you are analyzing, and generate your own highly innovative research idea.

"""